\pdfoutput=1

\documentclass[11pt]{article}

\usepackage[]{acl}

\usepackage{times}
\usepackage{latexsym}
\usepackage{hyperref}
 \hypersetup{
     colorlinks=true,
     urlcolor=black,
     }
\usepackage{caption}
\usepackage{subcaption}
\usepackage{graphicx}
\usepackage{float}

\usepackage[T1]{fontenc}

\usepackage[utf8]{inputenc}

\usepackage{microtype}

\title{Towards explainable evaluation of language models on the semantic similarity of visual concepts}

 \author{       Maria Lymperaiou,  {\bf George Manoliadis,}  {\bf Orfeas Menis Mastromichalakis,}  \\  {\bf Edmund G. Dervakos} and {\bf Giorgos Stamou} \\ Artificial Intelligence and Learning Systems Laboratory\\
School of Electrical and Computer Engineering \\
National Technical University of Athens \\
\{\texttt{\href{mailto:marialymp@islab.ntua.gr}{marialymp}}, \texttt{\href{mailto:eddiedervakos@islab.ntua.gr}{eddiedervakos}}\}\texttt{@islab.ntua.gr}, \texttt{\href{mailto:gmanoliad@mail.ntua.gr}{gmanoliad@mail.ntua.gr}}, \\ \texttt{\href{menorf@ails.ece.ntua.gr}{menorf@ails.ece.ntua.gr}}, \texttt{\href{gstam@cs.ntua.gr}{gstam@cs.ntua.gr}}
}

\begin{document}
\maketitle
\begin{abstract}
Recent breakthroughs in NLP research, such as the advent of Transformer models have indisputably contributed to major advancements in several tasks. However, few works research robustness and explainability issues of their evaluation strategies. In this work, we examine the behavior of high-performing pre-trained language models, focusing on the task of semantic similarity for visual vocabularies. First, we address the need for explainable evaluation metrics, necessary for understanding the conceptual quality of retrieved instances. Our proposed metrics provide valuable insights in local and global level, showcasing the inabilities of widely used approaches. Secondly, adversarial interventions on salient query semantics expose vulnerabilities of opaque metrics and highlight patterns in learned linguistic representations.  

\end{abstract}

\section{Introduction}
Semantic similarity between pairs of sentences serves a large variety of applications in the field of natural language processing, such as document retrieval, text classification, question answering and others. Even though such tasks have risen in popularity since the introduction of the Transformers \cite{transformer}, and despite the attention given on robustness and transparency of NLP transformers \cite{nlp_robustness, nlp_robustness2, nlp_transparency} few efforts have addressed explainable evaluation \cite{explainable-evaluation}. 

Text-Image retrieval is a real world semantic similarity application where the task is to feed a textual input to a system, and receive an image as a response. Visual details of the retrieved instance need to accurately correspond to the textual descriptions, often in a fine-grained fashion. Any mismatch between modalities can be easily perceived by humans, and captured by automated metrics. Such evident disagreements can act as starting points for further investigation, revealing inner processes on the semantic matching procedures. 

In this work, we aim to unveil the evaluation strategy of semantic similarity models. Specifically, we apply pre-trained transformers on visual vocabularies and obtain results via ranking. First, we address the shortcomings of traditional ranking metrics \cite{ir}, which provide either a binary answer (item found in top-k items or not), or position-informed variants (item found in the k-th position). However, such measures cannot provide detailed insights regarding the contribution of the scene constituents to the rank position. For example, if an instance is ranked in the k-th position, items in previous k-1 positions may be highly relevant to the ground truth one or on the contrary, highly irrelevant. To this end, we propose novel explainable ranking evaluation metrics that decompose and quantify the conceptual differences between ground truth and retrieved instances in local and global level. Even then, we observe that existing metrics lack a way to assess whether the top-ranked items are actually relevant to the query. For this reason, we construct adversarial queries where an attribute is replaced with a conceptually divergent one, in order to evaluate the response of a ranking system to distorted inputs. In all cases, frequently misperceived semantics captured by our evaluation framework reveal patterns imprinted in the learned representations of language models. Our overall approach is applicable regardless of the chosen language model or ranking system. 

\section{Related work}

A whole new world of possibilities in NLP has opened since the advent of the Transformer \cite{transformer}, with successful milestones such as BERT \cite{bert} and RoBERTa \cite{roberta} serving as backbone models for many applications. MPNet \cite{mpnet} combines permuted language modeling with masked language modeling to overcome the shortcomings of its predecessors. Towards reducing model sizes, knowledge distillation followed in DistilBERT/DistilRoBERTa \cite{distilbert}, MiniLM \cite{minilm} and TinyBERT \cite{tinybert}, as well as parameter reduction techniques implemented in ALBERT \cite{albert} achieve more compact models while maintaining performance. The textual semantic similarity task was greatly benefited by Sentence-BERT (SBERT) \cite{sbert}, a siamese-BERT variant that allows efficient embedding representations using the aforementioned models accordingly.

Traditional evaluation metrics such as HITS, Mean Reciprocal Rank (MRR), precision, recall and F-score \cite{ir} have dominated the field of information retrieval. While these metrics serve the purpose of assessing the retrieved information, they do not provide explainable means of justification. Explainable evaluation metrics \cite{explainable-evaluation} aim to address this challenge.

Lack of trust of neural methods due to biases, outdated training, and inaccurate assumptions has led to the need for explainable methods in language models. 
Research towards that direction has utilized Concept Attributions \cite{saietal, yuan2021}, Chunk Alignments \cite{magnolini}, Feature Importance \cite{rubino, treviso}, or Explanations by Simplification \cite{kaster}. Adversarial examples can also provide insights regarding the inner workings of obscure models, and are closely related to counterfactual explanations, placing them in the broader area of explainability \cite{linardatos2020explainable}. Numerous works address the problem of adversarial examples for natural language models \cite{10.1145/3374217}, with recent methods addressing the robustness of NLP models such as BERT through adversarial examples/attacks \cite{https://doi.org/10.48550/arxiv.1907.11932, https://doi.org/10.48550/arxiv.2004.09984}. In this work, we approach adversarial examples from a different perspective, first of all tackling a different problem than classification that most works do, and secondly by realizing the creation of adversarial examples on the semantic rather than the linguistic level, investigating the effect of semantic changes on the ranking of text-image retrieval systems.

\section{Overview}
\begin{figure*}[!h]
    \centering
        \includegraphics[width=0.88\textwidth]{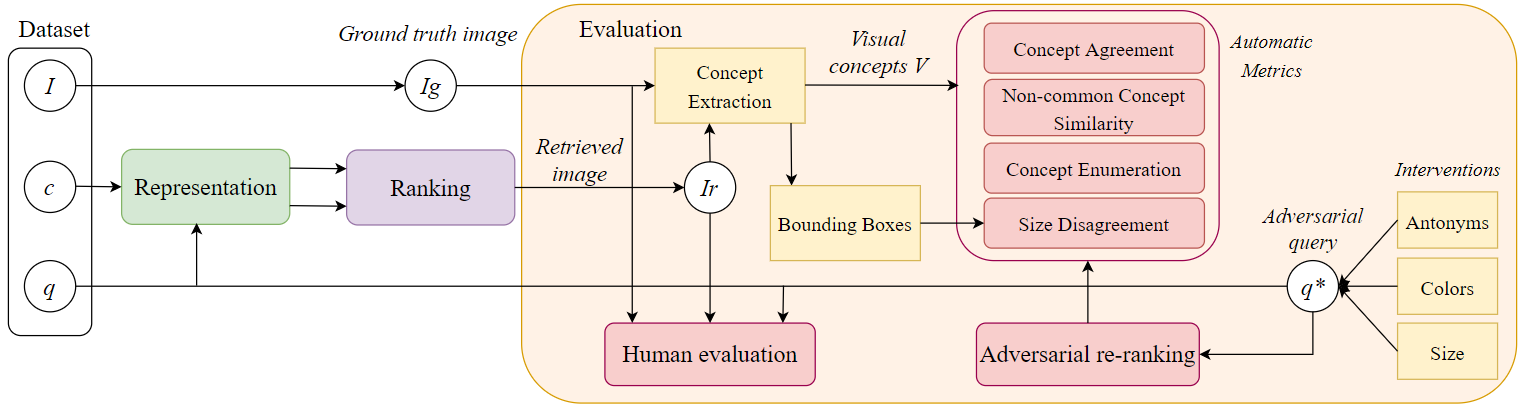}
        \caption{Overview of our workflow towards explainable evaluation.}
    \label{fig:overview}
\end{figure*}

\begin{figure}[!h]
    \centering
        \includegraphics[width=\columnwidth]{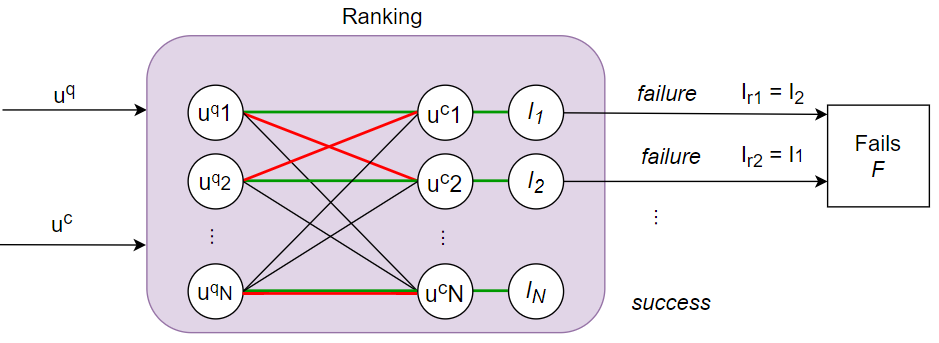}
        \caption{A closer look at ranking procedure. Green lines denote ground truth matchings, while red lines indicate matchings selected from maximum cosine similarity scores between query and corpus embeddings.}
    \label{fig:rank}
\end{figure}

Our workflow consists of three stages: \textbf{\textit{Representation}}, \textbf{\textit{ranking}} and \textbf{\textit{explainable evaluation}}. We view text-image retrieval as a query-corpus retrieval problem, exclusively exploiting linguistic information for representation and ranking, while revealing visual information only at the evaluation stage, where we compare retrieved images with the ground truth ones. As input, we consider a dataset of size $N$ that contains complex scene images $I_i\in\mathcal{I}$, accompanied by query-corpus pairs $(q_i, c_i), q_i\in\mathcal{Q}, c_i\in\mathcal{C}, i=1, 2, ..., N$ with each corpus $c_i$ consisting of an arbitrary number of sentences $s_j, j=1, 2, ..., l_c$. In the \textbf{\textit{representation}} stage, pre-trained sentence similarity transformers $M\in \mathcal{M}$ from SBERT embed $\mathcal{Q}, \mathcal{C}$ instances in a common vector space $U$. Cosine similarity scores between query-corpus embedding pairs in $U$ are sorted to provide a rank $R_i$ per query $q_i$ in the \textbf{\textit{ranking}} stage, with $R_i$ either lead to \textit{success}, if the ground truth image $I_{g_i}$ with corpus $c_i$ is returned at the top of the rank, or \textit{failure} otherwise. We provide a visual demonstration of the ranking procedure in Figure \ref{fig:rank}. All failures per model $M$, i.e. image pairs $(I_g, I_r)$ for which $I_g\neq I_r$  are stored in a set $\mathcal{F}$, which is further passed to the \textbf{\textit{evaluation}} stage. 
We then employ three methods to provide an understanding and evaluation towards failures: transparent ranking metrics (section \ref{sec:explain}), human evaluation and adversarial re-ranking (section \ref{sec:adversarial}). 
The overview of our proposal is presented in Figure \ref{fig:overview}.

\subsection{Visual concepts in language} 
Visual vocabularies contain descriptions about real life scenes, including objects, relationships and attributes. Datasets that connect visual vocabularies paired with images, such as Visual Genome \cite{visualgenome}, COCO \cite{coco} and Flickr \cite{flickr} set our sources to construct purely textual query-corpus pairs, assuming that necessary visual information is contained within the high quality annotations of those datasets. In particular, the annotation diversity allows either shorter, global descriptions, as in Flickr and COCO captions, or detailed descriptions in local level, as in Visual Genome region descriptions, concatenated in a corpus $c_i$ per image $I_i$. 

\subsection{Optimal embedding representation} 
Obtaining an overall representation of a corpus $c_i$ is not trivial, as existing transformers can handle up to a certain number of input tokens per sentence. To resolve this, we can independently embed each corpus sentence $s_j\in c_i, j=1, 2, ..., l_c$ using a model $M\in \mathcal{M}$, and then calculate the average of all vectors $v_j^c$. Therefore,  $u_i^c = \frac{1}{l_c} \sum_{j=1}^{l_c}v_j^c \in U$ serves as the averaged representation for $c_i$.
Another approach is to leverage state-of-the-art abstractive summarizers \cite{pegasus, t5} to obtain a meaningful shorter version of $c_i$ while maintaining semantics as much as possible, and then apply $M\in \mathcal{M}$ only once per $c_i$. 
Query representations $u_i^c\in U$ are produced by inserting each $q_i\in\mathcal{Q}$ in a model $M\in\mathcal{M}$, or by averaging over representations when $q_i$ comprises from more than one sentences.

\subsection{Ranking} 
Given a model $M$, each query representation $u_i^q\in U$ is paired with all corpus representations $\{u_1^c, u_2^c, ..., u_N^c\}\in U$, and cosine similarity scores 
are calculated for each pair. Higher cosine similarity scores yield more similar representations, therefore sorting from higher to lower scores provides the ranking $R_i$ per $q_i$. The process is repeated for all $N$ images resulting in $N^2$ calculations. 

Traditional metrics evaluate the ranking success, coarsely indicating the representation quality of each $M\in \mathcal{M}$. Recall@k returns the proportion of ground truth images found in top-k ranked instances for all queries $q_1, q_2,...q_N$, given that each $q_i$ has only one ground truth $c_i$. Mean Reciprocal Rank (MRR) is the averaged of the inverse of the ground truth rank position $\mathsf{rank}_i$ for each $c_i$ given $q_i$, considering the top-k items: MRR@k = $\frac{1}{N}\sum_{i=1}^{N}\frac{1}{\mathsf{rank}_i}$ for each $\mathsf{rank}_i\geq k$. We calculate Recall@k and MRR@k for k=5, 10, $N$. Also, we calculate the median rank position for all $c_i$.

\section{Explaining failures} 
\label{sec:explain}
We count as failure $f_i = (I_g, I_r)_i \in \mathcal{F}$ any instance of a ground truth image $I_{g_i}$ with corpus $c_i$ that was not ranked in the first position ($\mathsf{rank}_i \neq 1$) given $q_i$; instead another image $I_{r_i} \neq I_{g_i}$ with $c_r \neq c_i$ achieved $\mathsf{rank}_r=1$. Following the 'blind' evaluation strategy of traditional ranking metrics, we provide a measure of retrieval failures as the cardinality of the failure set: $F=|\mathcal{F}|$ for each $M$.

However, it is not possible to verify if $I_{r_i}$ can accurately satisfy $q_i$ without exploiting visual information. To this end, we exploit visual annotations and human perception to quantify the suitability of each $I_{r_i} \in f_i$ with respect to $q_i$. By decomposing all semantics that contribute to the suitability of each $I_{r_i}$ we obtain a discrete and transparent conceptual measure of similarity between $(I_g, I_r)_i$.

\subsection{Towards explainable evaluation metrics} 
We design four evaluation stages for all failures $f_i$, starting from more influential concepts and moving towards less prevalent details. Visual concepts are focused on scene \textbf{\textit{objects}}. For fair comparison with traditional ranking metrics, we demonstrate a \textit{query-agnostic} evaluation approach: we compare concepts between retrieved and ground truth images without considering query semantics. In the next paragraphs we drop $i$ subscript for simplicity.
\paragraph{Concept agreement - CA} Considering $\mathcal{V}$ as a set of visual concepts, \textbf{concept agreement}
measures the percentage of ground truth concepts $\mathcal{V}(I_g)$ contained in the retrieved concept set $\mathcal{V}(I_r)$ over all $\mathcal{V}(I_g)$ concepts for each $f_i$. Let $V_{(g, r)} = \mathcal{V}(I_g) \: \cap \: \mathcal{V}(I_r)$  the set of common concepts:
\begin{center}
    $CA_{f} = \frac{|V_{(g, r)}|}{|\mathcal{V}(I_g)|}$, $f = (I_g, I_r)$ \\
\end{center}

Higher \textbf{\textit{CA}} indicates higher concept similarity. For example, if $\mathcal{V}(I_r)=\{\mathsf{Dog, Frisbee, Park}\}$ and $\mathcal{V}(I_g)=\{\mathsf{Dog, Ball, Park}\}$, then the \textbf{\textit{CA}}=$\frac{2}{3}$. On the other hand, if $\mathcal{V}'(I_r)=\{\mathsf{Cat, Fish}\}$, then  \textbf{\textit{CA}}=0, as no overlap exists. This way we can confidently conclude that the first retrieved image is conceptually closer to the ground truth than the second, and by extension the model used to retrieve the first image is better with respect to \textbf{\textit{CA}}.

\paragraph{Non-common concept similarity - NCS} aims to provide a distance measure between concepts present exclusively in either $\mathcal{V}(I_g)$ or $\mathcal{V}(I_r)$. For example, we would expect the set $\{\mathsf{Dog,Frisbee,Park}\}$ to be more similar to $\{\mathsf{Dog,Ball,Park}\}$ than $\{\mathsf{Dog,Cat,Park}\}$, since the non-common concept $\mathsf{Frisbee}$ is conceptually closer to $\mathsf{Ball}$ than $\mathsf{Cat}$. Mathematically, let  $D_g = V_{(g, r)} - \mathcal{V}(I_r)$ and $D_r = V_{(g, r)} - \mathcal{V}(I_g)$, with both $D_g, D_r\neq \emptyset$. Other than that, $D_g$ and $D_r$ may contain different number of concepts. Then, a measure of concept distance can be provided by calculating the path similarity score $ps$ of corresponding WordNet \cite{wordnet} synset pairs, based on the shortest available path between those two concepts. Path similarity $ps$ ranges between 0 and 1. 

An \textit{optimistic} \textbf{\textit{NCS}} metric returns the maximum possible cumulative $ps$ averaged over the number of pairs, by appropriately selecting concept pairs between non-empty $D_g$ and $D_r$. The maximization of \textbf{\textit{NCS}} requires a dynamic programming solution, as naive strategies taking into account all possible $D_g$ and $D_r$ pairs would yield a factorial amount of combinations. To trespass this prohibitive complexity, we create a bipartite graph $G=(D_g, D_r, E)$ from $D_g$ and $D_r$: all concept nodes from the one set are matched with all the nodes of the other via edges $e_y \in E, y=1, 2, ..., |D_g|\times |D_r|$, while no edges are allowed within the same set. Edge weights $w_{e_y}$ correspond to WordNet $ps$ scores between synsets of connected nodes.

Consequently, the \textit{maximum weight bipartite matching} on $G$ refers to pairing $D_g$ and $D_r$ concepts so that the cumulative edge weight is maximized. An optimized version of the Hungarian algorithm \cite{hungarian_algo, max_weight_matching} implemented by NetworkX\footnote{\href{https://networkx.org/documentation/stable/reference/algorithms/generated/networkx.algorithms.matching.max_weight_matching.html}{NetworkX max weight matching}} reduces the computational complexity of finding the maximum $ps$ to $O(|V|^ 3)$, where $|V|= max(|D_g|, |D_r|)$.

Therefore, \textbf{\textit{NCS}} can be written as:
\begin{center}
    $NCS_{f} = avg(max\_weight\_match(G))$, $G=(V_{(g, r)} - \mathcal{V}(I_r), V_{(g, r)} - \mathcal{V}(I_g), E))$
\end{center}

Higher \textbf{\textit{NCS}} scores reveal more similar concepts. 

\paragraph{Concept enumeration - CE} Real world scenes may contain repeated instances of same-class concepts, forming concept multisets $\mathcal{V}_m=\{(\mathcal{V}_1, |\mathcal{V}_1|), (\mathcal{V}_2, |\mathcal{V}_2|), ..., (\mathcal{V}_x, |\mathcal{V}_x|) \}$, where $\mathcal{V}_1, \mathcal{V}_2, ..., \mathcal{V}_x$ denote concept categories, and $|\mathcal{V}_1|, |\mathcal{V}_2|, ..., |\mathcal{V}_x|$ cardinalities per category. The cardinality per concept category is called concept multiplicity in the multiset. \textbf{\textit{CE}} penalizes differences in multiplicities between common concepts of $I_g$ and $I_r$ for each $f_i$:

\[CE_{f} = \sum_{j=1}^{x}||\mathcal{V}_j(I_g)| - |\mathcal{V}_j(I_r)||_{\mathcal{V}_j(I_g)=\mathcal{V}_j(I_r)}\]

Higher \textbf{\textit{CE}} scores demonstrate higher enumeration disagreement, deeming lower \textbf{\textit{CE}} values more favorable. For example, if $I_g$ contained 10 dogs and 1 frisbee $\{(\mathsf{Dog},10), (\mathsf{Frisbee},1)\}$, a retrieved $I_r$ with 1 dog and 1 frisbee would have \textbf{\textit{CE}}=9, while an $I'_r$ with 10 dogs and 1 ball would have a \textbf{\textit{CE}}=0. Therefore, the first image yields a worse CE score than the second, even though the second would have worse \textbf{\textit{CA}} and \textbf{\textit{NCS}} scores than the first one. 

\paragraph{Size disagreement - SD} Even in cases where there is a high agreement of objects and multiplicities between $I_g$ and $I_r$, disagreement in object sizes may correspond to semantically divergent scenes. For example an image with a dog in the foreground (large bounding box) is different than an image of a dog in the background (small bounding box). To capture this difference, we design an \textit{optimistic} \textbf{\textit{SD}} metric which returns the area differences of bounding boxes $\mathcal{D}_A = |A_g - A_r|$ for all available object matchings. Such matchings occur by pairing concepts of the same category $u$ between $\mathcal{I}_g$ and $\mathcal{I}_r$ up to the point that no more unique pairs can be constructed. This is equivalent of creating bipartite graph $G_u=(\mathcal{V}_u(I_g), \mathcal{V}_u(I_r), E)$, where $\mathcal{V}_u(I_g), \mathcal{V}_u(I_r)$ belong in the same $u$ and edge weights $w_{e_y}, e_y \in E, y=1, 2, ..., |\mathcal{V}(I_g)|\times |\mathcal{V}(I_r)|$ denote the area difference $\mathcal{D}_A$ between concept nodes. Pairing concepts with similar bounding box areas can be considered as the optimal choice, therefore node pairs connected by lower edge weights  $w_{e_y}$ are preferred. Finding the \textit{minimum weight matching} provides the most similar pairs size-wise, and can be solved in polynomial time using the NetworkX\footnote{\href{https://networkx.org/documentation/stable/reference/algorithms/generated/networkx.algorithms.bipartite.matching.minimum_weight_full_matching.html}{NetworkX min weight full matching}} implementation of Karp algorithm \cite{min_weight_matching}. The matching process is repeated for all concept categories in the multiset $\mathcal{V}_m$, resulting in a set of graphs $G_m$:

\begin{center}
    \[SD_{f} = \sum^{\mathcal{V}_u \in \mathcal{V}_m} avg(min\_weight\_match(G_u)), \]
    \[G_u=(\mathcal{V}_u(I_g), \mathcal{V}_u(I_r), E), G_u \in G_m \]
\end{center}

A simplified \textit{binary} version of SD increases a sum if area differences of paired concepts are above a predefined threshold $T_D$.

\subsection{Human evaluation via crowdsourcing}
\textit{Query-agnostic} evaluation regards all scene semantics, even if in fact they are not present in the query. On the other hand, incorporating query information at evaluation stage conditions concept importance upon the presence of a concept in the query, forming a \textit{query-informed} evaluation strategy. We conducted \textit{query-informed} human evaluation experiments considering all failures in $\mathcal{F}$ and penalizing semantic disagreements only if those semantics are mentioned in $q_i$. Evaluators were primarily asked to mark which salient semantics were clearly misinterpreted in retrieved images with respect to the given query among the options: \textit{object class, object color, object enumeration, action, size, details}. Otherwise, if $I_{r_i}$ can be considered as conceptually similar to $I_{g_i}$, it is marked as \textit{successful alternative}.
Additionally, the overall retrieval quality is cross checked via qualitative ratings, assessing the conceptual similarity between $I_{g_i}$, $I_{r_i}$ given $q_i$. Despite being unfair to compare with the -stricter- automated metrics, we expect lower values for \textit{object enumeration} and \textit{size} failure classes comparing to \textbf{\textit{CE, SD}} metrics.

The crowdsourcing experiment reveals the most frequently misinterpreted attributes or combinations of attributes. Loss of conceptual information can be either attributed to dataset quality, i.e. salient query semantics not present in corpus, or on the capacity of the linguistic representations. Keyword matching between $q_i$ and $c_i$ excludes cases where the ground truth query-corpus pair contains very few common concepts, enabling the remaining samples to reveal patterns within the learned representations.

\section{Adversarial re-ranking}
\label{sec:adversarial}
\begin{figure}[!h]
    \centering
        \includegraphics[width=0.97\columnwidth]{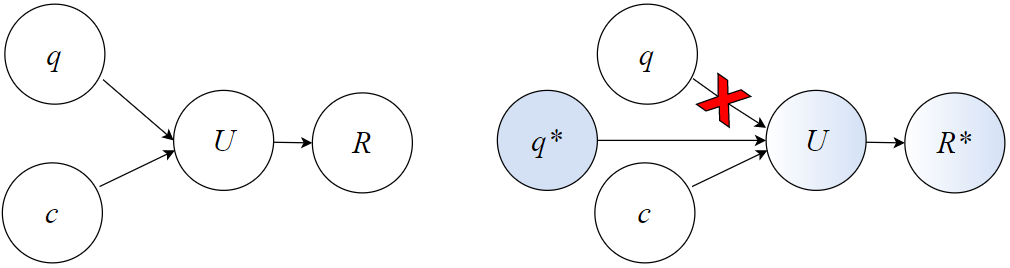}
        \caption{Conventional causal graph (left) and adversarial intervention causal graph (right) when $q\rightarrow q^*$.}
    \label{fig:graph}
\end{figure}

We create adversarial queries $q\rightarrow q^*$ targeting key attributes and produce respective representations in $U$, upon which adversarial rankings $R^*$ per $q^*$ are extracted. Figure \ref{fig:graph} provides the causal graph of adversarial interventions for any $q_i\in \mathcal{Q}$.
\subsection{Substituting salient attributes} 
We perturb salient semantics in queries $q_i \in \mathcal{Q}$, producing $q^*_i \in \mathcal{Q}^*$, and evaluate the changes occurring in the rank. An appropriate non-minimal adversarial perturbation must conceptually reverse salient semantics, be focused on an individual semantic each time, and the resulting query $q^*_i$ should be linguistically correct. With respect to those requirements and in order to restrict the search space of adversarials, we target substituting \textbf{\textit{object attributes}}. Initially, generic adversarial queries include replacing attributes with their antonyms. More refined subsequent adversarials focus on replacing object colors and sizes; such substitutions are discrete, fast and controllable.
\paragraph{Antonyms} are extracted via relevant WordNet functions for any adjective present in a query. If more than one antonyms are returned, one is randomly picked to substitute the actual word.
\paragraph{Color substitution} refers to changing colors present in the sentence with another distant color. Color distance is provided via the RGB values of Matplotlib colors\footnote{\href{https://matplotlib.org/3.3.3/gallery/color/named_colors.html}{Matplotlib colors}}. We set a proximity threshold to ensure perceptually non-negligible color changes. Two possible substitutions are attempted: either considering all RGB colors (\textit{color-all)}, or colors only mentioned in the dataset (\textit{color-in)}.
\paragraph{Size substitution} is an antonym substitution specialized in sizes. Words such as \textit{large, big, enormous, huge} are substituted with a random choice among \textit{small, little, minor, tiny} and vice versa.

\subsection{Re-ranking evaluation} 
Adversarial query representations $u^{q*}_i\in U$ of $q^*\neq q$ with $u^c_i\in U$ of corpus $c_i$ may directly influence the final ranking $R*$ when $\mathsf{rank}^*_i$< $\mathsf{rank}_i$ or inversely $\mathsf{rank}^*_i$>$\mathsf{rank}_i$. 
Intuitively, any non-negligible perturbation of $q$ should result in worse position $\mathsf{rank}^*_i$>$\mathsf{rank}_i$, as the adversarial query representation $u^{q*}_i$ would diverge from $c_i$ comparing to  $u^{q}_i$, due to the substitution of the actual semantic with a conceptually different one. However, given the relative nature of ranking, some instances may stay in the same position $\mathsf{rank}^*_i$=$\mathsf{rank}_i$, or even go higher. Ascending in the rank does not imply a better $q^*_i, c_i$ matching, except if their in between cosine similarity increases; instead, the distorted representations push lower previously higher-ranked instances, virtually improving some $\mathsf{rank}^*_i$. In any case, we expect all ranking metrics to perceptibly drop, as we pull apart ground truth matchings in $U$. 

\section{Experiments}

We implement the same experimental strategy on Flickr8k, Flickr30k and VG$\,\cap\,$COCO datasets, obtaining query-corpus representations in a common semantic space $U$ for a variety of models $M\in \mathcal{M}$. 
The code for our experiments is provided in submitted supplementary material.
\subsection{Ranking results} 
We focus on presenting experiments on VG$\,\cap\,$COCO of $N$=34k images, which is the most challenging: the dataset size $N$ itself, as well as the more detailed region descriptions of Visual Genome which comprise a larger corpus set $\mathcal{C}$, require accurate linguistic representations in order to retrieve more relevant images.  Table \ref{tab:rank-vgcoco} presents a subset of ranking results on VG$\,\cap\,$COCO.

\begin{table*}[h!]
\centering
\begin{tabular}{l|p{25pt}p{25pt}p{25pt}|p{25pt}p{25pt}p{25pt}|c|c}
\hline
Name     &  & Recall(\%)$\uparrow$ &  &  & MRR(\%)$\uparrow$ &   & Median & Fails \\
     & $@1$ & $@5$ & $@10$ & $@5$ & $@10$ &  $@all$ & Rank $\downarrow$ & (\%) $\downarrow$  \\
\hline
all MiniLM L12 & 13.31 & 26.93 & 34.31 & 18.18 & 19.16 & 20.45 & 34 & 86.69 \\
paraphrase MiniLM L12 & 14.24 & 29.89 & 38.34 & 19.89 & 21.02 & 22.37 & 24 & 85.76 \\
paraphrase MiniLM L3 & \textbf{15.31} & \textbf{30.92} & \textbf{39.48} & \textbf{20.97} & \textbf{22.11} & \textbf{23.42} & 23 & \textbf{84.69} \\
paraphrase MiniLM L6 & 14.39 & 30.01 & 38.51 & 19.96 & 21.10 & 22.46 & 24 & 85.61 \\
paraphrase TinyBERT L6 & 14.55 & 30.38 & 39.12 & 20.27 & 21.43 & 22.82 & \textbf{22} & 85.45 \\
paraphrase albert base & 13.12 & 27.83 & 36.11 & 18.43 & 19.54 & 20.91 & 28 & 86.88 \\
paraphrase albert small & 14.56 & 30.25 & 39.04 & 20.23 & 21.40 & 22.75 & 23 & 85.44 \\
paraphrase distilroberta & 14.51 & 30.39 & 38.91 & 20.23 & 21.36 & 22.74 & \textbf{22} & 85.49 \\
paraphrase mpnet & 13.99 & 28.99 & 37.40 & 19.39 & 20.50 & 21.84 & 26 & 86.01 \\
stsb distilroberta base & 13.41 & 27.04 & 34.28 & 18.32 & 19.28 & 20.55 & 35 & 86.59 \\
stsb mpnet base & 14.05 & 28.26 & 35.93 & 19.23 & 20.24 & 21.49 & 32 & 85.95 \\
stsb roberta base & 13.69 & 27.38 & 34.79 & 18.67 & 19.65 & 20.92 & 34 & 86.31 \\
\hline
\end{tabular}
\caption{Rank results on the VG$\,\cap\,$COCO dataset for our best 12 models. Full Table (\ref{tab:vgcoco}) in Appendix.}
\label{tab:rank-vgcoco}
\end{table*}
\subsection{Optimal model choice} 
Selected language models are designed for semantic similarity, and according to the datasets they have been pre-trained on, they can be divided in: \textit{all-} models pretrained and fine-tuned on 1B sentence pairs from multiple sources;  \textit{multi-qa-} trained on 215M diverse question-answer pairs, learning to map queries to passages; \textit{stsb-} models trained on the STSbenchmark, which contains sentence pairs annotated with similarity scores; \textit{paraphrase-} models with more than 86M paraphrase sentence pairs containing more challenging and uncurated characteristics comparing to STSb; \textit{nq-} models trained with 100k real Google search queries mapped to Wikipedia passages; \textit{nli-} models incorporate natural language inference data pairs (premise/hypothesis), included in AllNLI dataset.

By conducting a large number of experiments to estimate the performance of such models on visual vocabularies, we observe certain patterns in ranking results. In all experiments, most paraphrase models consistently outperform the rest. Paraphrasers have been pre-trained on image captions (COCO, Flickr), which actually serve as paraphrasing data: during the construction of these datasets, annotators have independently produced varying descriptions for the same concepts. Query-corpus pairs can be viewed as the one being a paraphrase of the other, thus paraphrasers have learned a suitable representation for this matching, together with their exposure to visual vocabularies.

\subsection{Explainable evaluation} 
\begin{figure}[h!]
\begin{subfigure}[b]{0.47\columnwidth}
\centering 
\includegraphics[width=\columnwidth, height=2.5cm]{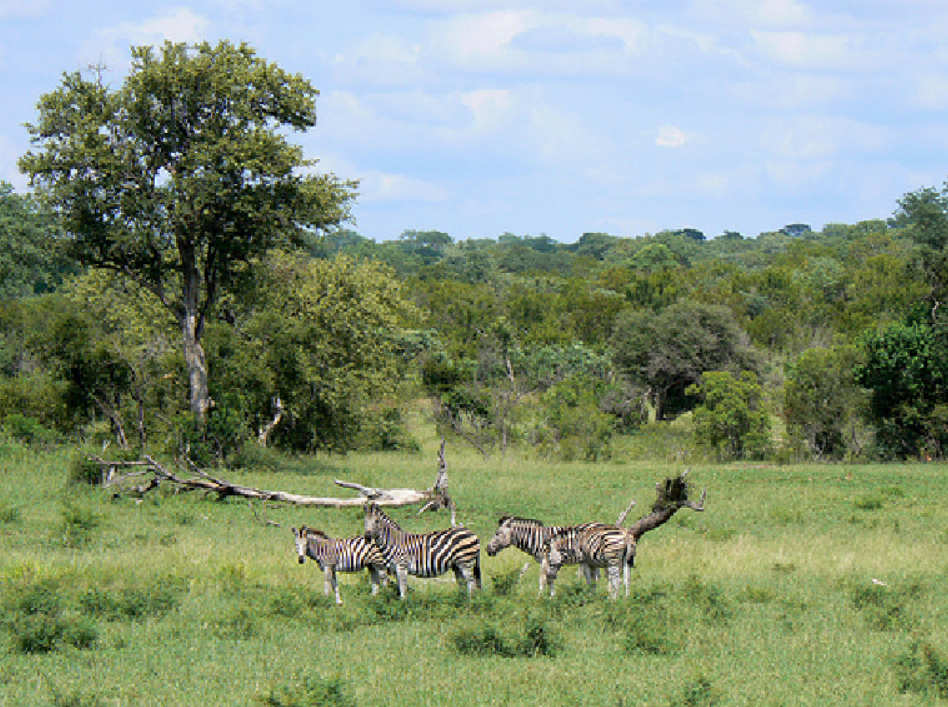}
\end{subfigure}\hfill
\begin{subfigure}[b]{0.47\columnwidth}
\centering 
\includegraphics[width=\columnwidth,height=2.5cm]{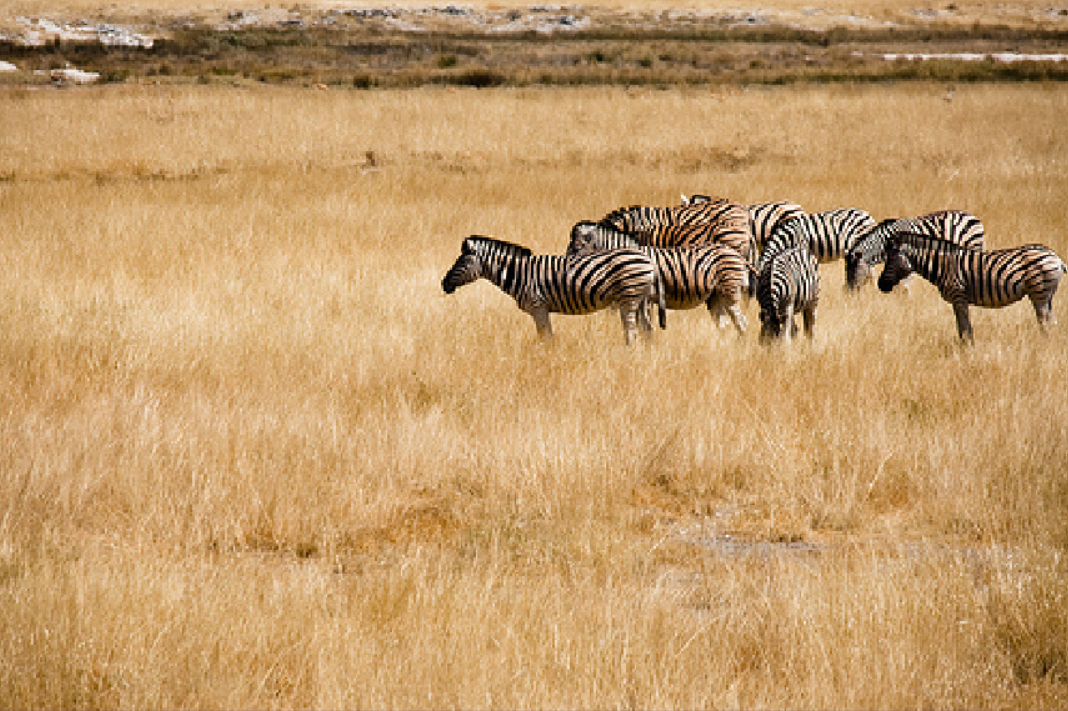}
\end{subfigure}
\caption{A herd of zebras grazing in a lush green field}
\label{fig:example}
\end{figure}

In all following experiments, we consider results from the best performing model MiniLM-L3 on VG$\,\cap\,$COCO.
In total, $F$=28817 queries failed to retrieve their corresponding ground truth $I_g$.
\paragraph{Local evaluation}
The real power of our proposed metrics lies in local level.
We present an example from the \textit{color} and \textit{details} failure category below. Given a query $q_i$ (caption), Figure \ref{fig:example} shows the ground truth $I_{g_i}$ (left) and the retrieved $I_{r_i}$ (right). 

The set of ground truth object synsets is \{$\mathsf{trunk.n.01}$, $\mathsf{hill.n.01}$, $\mathsf{tree.n.01}$, $\mathsf{sky.n.01}$, $\mathsf{field.n.01}$, $\mathsf{branch.n.01}$, $\mathsf{head.n.01}$, $\mathsf{leg.n.01}$, $\mathsf{leaf.n.01}$, $\mathsf{zebra.n.01}$, $\mathsf{mane.n.01}$\} of cardinality 11, and the set of retrieved ones is \{$\mathsf{grassland.n.01}$, $\mathsf{field.n.01}$, $\mathsf{zebra.n.01}$, $\mathsf{mane.n.01}$, $\mathsf{grass.n.01}$\} of cardinality 5. Common synsets are \{$\mathsf{zebra.n.01}$, $\mathsf{mane.n.01}$, $\mathsf{field.n.01}$\} of cardinality=3, resulting in \textbf{\textit{CA}}$_i$=27.28\%.
Regarding \textbf{\textit{NCS}}, the constructed bipartite graph $G$ contains $|V|$=10, $|E|$=16, and the best matched synset pairs according to the \textit{maximum weight matching} are \{$\mathsf{hill.n.01}$, $\mathsf{grassland.n.01}$\} with $ps$=0.111, and \{$\mathsf{tree.n.01}$, $\mathsf{grass.n.01}$\} with $ps$=0.167. The average $ps$ for all matched pairs leads to \textbf{\textit{NCS}}$_i$=0.139.
Common object enumeration provides the following multisets: $\mathcal{V}^g_m$ = \{$\mathsf{zebra.n.01}$, 5, $\mathsf{field.n.01}$, 1, $\mathsf{mane.n.01}$, 1\} and $\mathcal{V}^r_m$ =\{$\mathsf{zebra.n.01}$, 7, $\mathsf{field.n.01}$, 1, $\mathsf{mane.n.01}$, 5\}. Therefore, \textbf{\textit{CE}}$_i$= 6.
As for \textbf{\textit{SD}} for $T_D$=1, 3 bipartite graphs are created for the 3 common synsets. The first graph $G_{mane}$ contains $|V|$=6 and $|E|$=5, resulting in 1 \textit{minimum weight matching} of weight $D_A$=2.30$\geq T_D$. Therefore \textbf{\textit{SD}}$_{mane}$=1=\textbf{\textit{SD}}$_i$. The second graph $G_{zebra}$ consists of $|V|$=11 and $|E|$=28, resulting in 4 \textit{minimum weight matchings}, from which none trespassed the threshold $T_D$, resulting in \textbf{\textit{SD}}$_{zebra}$=0, thus maintaining \textbf{\textit{SD}}$_i$=1. Finally, the $G_{field}$ graph of $|V|$=2 and $|E|$=1, leads to 1 \textit{minimum weight matching} of weight $D_A$=1.369$\geq T_D$, resulting in \textbf{\textit{SD}}$_{field}$=1, which increases the total sum \textbf{\textit{SD}}$_i$=2. Having in total 6 matches for all three graphs, the \textbf{\textit{averaged SD}}$_i$=33.3\% for this $f_i$.

Perceptually, a major $I_{g_i}, I_{r_i}$ disagreement can be attributed to not satisfying \textit{lush, green} attributes rather than semantics addressed by our metrics. Indeed, human evaluators rated $I_{r_i}$-$q_i$ relevance with 6/10 on average and all of them marked \textit{details} and \textit{color} as the failure categories. As for traditional ranking metrics, $I_{g_i}$ was placed in $\mathsf{rank}_{g_i}$=294 with reciprocal rank score of 0.0034 and R@k=0, k=1,5,10. Obviously, we cannot extract much information in local level about how much $I_{g_i}$ and $I_{r_i}$ conceptually deviate and what we should potentially regard and request from retrieved instances (\textit{colors} such as \textit{green} instead of \textit{yellow}, \textit{details} such as \textit{lush} instead of \textit{arid}) to ascend in the rank. To this end, we conclude that traditional ranking metrics are only helpful in a very abstract level.

\paragraph{Human evaluation} 

Results regarding misperceived semantics classes are presented in Table \ref{tab:human}. The 82.52\% of evaluated image pairs resulted in one semantic class disagreement, while the remaining 14.56\% and 2.91\% contained two and three semantic class disagreements respectively. The average rating over all classes was 8.47/10.

First, human evaluation experiments can indicate the degree of strictness of our automated metrics, as any \textit{query-agnostic} metric may over-penalize semantics present in the $I_{g_i}$ and $c_i$ but not in $q_i$. Indeed, \textit{query-informed} variants of our metrics are more relaxed. Moreover, patterns in reported failures also indicate patterns imprinted in the learned linguistic representations. Traditional ranking metrics cannot derive such fine-grained observations.

\begin{table}[h!]
\centering
\begin{tabular}{p{36pt}p{14pt}p{15pt}p{17pt}p{15pt}p{20pt}p{20pt}}
\hline
Altern- &       &  Obj. &  \%   &  & Action &  Detail \\
atives \%  & class &  color & enum. & size & \hfil\%   & \hfil\% \\
\hline
23.30 & 5.83  & 17.48 & 11.65 & 6.14 & 7.77 & 54.37 \\
\hline
\end{tabular}
\caption{Semantics disagreement percentage per class}
\label{tab:human}
\end{table}

\paragraph{Rules in failures}
The frequent class of \textit{successful alternatives} indicates that even when automatic metrics consider an $I_{r_i}$ as failure, it may actually be a conceptually correct answer to $q_i$.  Qualitative analysis over \textit{successful alternatives} further demonstrated that almost all $(I_{g_i}, I_{r_i})$ pairs of this class were visually divergent, even though conceptually equivalent. Also, \textit{details} and \textit{object color} failure classes appeared often enough, indicating that those semantics are rather bypassed in order for others to be preserved. Combinations of semantics did not present any significant pattern; all semantics co-occurrences appeared in less than 10\% of the evaluated instances. However, we did observe some frequent rules, which can be translated as: \textit{if semantic A disagrees, then semantic B will disagree as well}. The rule action$\rightarrow$details (\textit{if action appears then details will appear}) is observed in 54.37\% of the instances containing \textit{action}; object enumeration $\rightarrow$object color covers 17.48\% of the instances containing \textit{color}; finally, the reverse rule object color$\rightarrow$object enumeration was observed in 11.65\% of the instances containing \textit{enumeration}.

\paragraph{Global evaluation}
We present global \textit{query-agnostic} results for our metrics. Despite our metrics being more meaningful in local level, global evaluation is useful for model benchmarking.

With 134630 common concepts between all $(I_g, I_r)_i\in \mathcal{F}$, the \textbf{\textit{average concept agreement (CA)}} value is 22.29\%, meaning that on average almost the 1/4th of $I_g$ concepts appear in $I_r$.

With 903987 non-common concepts between all $(I_g, I_r)$, and 134630 common ones, we retrieve 627833 and 110839 WordNet synsets respectively. The \textit{maximum weight matching} between non-common synsets results in 184747 maximum weight matchings, equivalent to the 29.43\% of all non-common synsets. Averaging over matchings (WordNet path similarities) for all $(I_g, I_r)$, provides the \textbf{\textit{average non-common concept similarity (NCS)}} score of 0.122.

With 41244 concept sets of same multiplicity and 69595 of different multiplicities regarding matched concept categories for all $(I_g, I_r)$, \textbf{\textit{most common concept enumeration (CE)}}=1 and \textbf{\textit{average CE}}=8.638 instances for concepts of the same category reveals that in most cases there are not major enumeration differences.

Focusing on the binary \textbf{\textit{SD}}, we set the area difference threshold $T_D$=100\%, increasing \textbf{\textit{size disagreement (SD)}} by 1 iff $\mathcal{D}_A \geq 1$ between two concept bounding box areas. Thus, \textbf{\textit{average SD}}=20.35\% for all $(I_g, I_r)$, indicating that around 1/5th of common objects have non-negligible size differences.

Our metrics in global level reveal some extra capabilities. Most lower-ranked instances contained erroneous annotations, allowing a \textit{post-hoc dataset cleaning} step that could not have been automatically realized otherwise. 

Global results regarding our proposed metrics for all the models are presented in Table \ref{tab:our_metrics}. Moreover, we offer some additional insights:
\begin{itemize}
    \item Object hit: total number of common objects found between ground truth - wrongly retrieved images ($I_g, I_r$) at top-1 position.
    \item Object miss: total number of ground truth objects not found in top-1 retrieved images.
    \item Matched \% synsets: Percentage of ground truth synsets found in top-1 retrieved images out of all ground truth synsets.
    \item Average \% object enumeration disagreement: percentage of objects having the wrong number of instances between ground truth and top-1 retrieved over all ground truth objects (both having right or wrong number of instances).
\end{itemize}
As observed, the various explainable metrics indicate different models as best/worst performers, revealing that fine-grained evaluation may disagree with traditional coarse evaluation, while providing some useful insights. 

\begin{table*}[h!]
\hspace{-1cm}
\begin{tabular}{c|cccc|cccc}
Name & CA & NCS & CE & SD  & obj & obj & matched \%  & avg\% enum   \\
& $\uparrow$ & $\uparrow$ & $\downarrow$ & $\downarrow$ & hit $\uparrow$ & miss $\downarrow$ & synsets $\uparrow$ &  disagr.$\downarrow$ \\
\hline
all distilroberta & 21.75 & 0.12 & 9.14 & 20.02 &  136025 & 932261 & 29.43 & 2.40 \\
all MiniLM L12 & 21.72 & 0.12 & 8.88 & 19.88  & 134188 & 920214 & 29.23 & 2.35 \\
all MiniLM L6 & 21.90 & 0.12 & 8.91 & 19.53  & 135687 & 926717 & 29.27 & 2.39 \\
all mpnet base & 21.74 & 0.12 & 9.14 & 19.24  & 135278 & 926724 & 29.44 & 2.39 \\
all roberta large & 21.85 & 0.12 & 9.13 & 19.60  & 136720 & 935614 & 29.42 & 2.39 \\
multi qa distilbert cos & 21.84 & \textbf{0.13} & 9.06 & 16.97  & 140754 & 1007800 & \textbf{30.60} & 2.49 \\
multi qa MiniLM L6 cos & 21.93 & \textbf{0.13} & 8.69 & 18.86  & \textbf{141399} & 1008889 & 30.21 & 2.45 \\
multi qa mpnet base cos & 21.87 & \textbf{0.13} & 9.24 & \textbf{16.69}  & 140455 & 1004153 & 30.54 & 2.52 \\
nli distilroberta base & 22.12 & 0.12 & 8.89 & 19.68  & 138882 & 943947 & 29.75 & 2.45 \\
nq distilbert base & 21.39 & 0.12 & 9.03 & 17.53  & 139608 & 1020776 & 30.27 & 2.38 \\
paraphrase albert base & 22.13 & 0.12 & 8.92 & 19.71  & 136867 & 927060 & 29.52 & 2.43 \\
paraphrase albert small & \textbf{22.40} & 0.12 & 8.89 & 19.48  & 136319 & 909759 & 29.69 & 2.47 \\
paraphrase distilroberta & 22.28 & 0.12 & 8.63 & 19.49  & 135679 & 911325 & 29.32 & 2.39 \\
paraphrase MiniLM L12 & 22.28 & 0.12 & 8.69 & 20.05  & 136102 & 910631 & 29.23 & 2.39 \\
paraphrase MiniLM L3 & 22.29 & 0.12 & 8.64 & 20.35  & 134630 & \textbf{903987} & 29.43 & 2.42 \\
paraphrase MiniLM L6 & 22.07 & 0.12 & 8.61 & 19.37  & 134623 & 914535 & 29.46 & 2.37 \\
paraphrase mpnet & 22.26 & 0.12 & 8.71 & 19.46  & 136482 & 911716 & 29.08 & 2.39 \\
paraphrase TinyBERT L6 & 22.15 & 0.12 & 8.55 & 19.67  & 134808 & 916573 & 29.17 & 2.36 \\
stsb distilroberta base & 22.09 & 0.12 & 8.91 & 19.99  & 136181 & 928388 & 29.81 & 2.45 \\
stsb mpnet base & 21.84 & 0.12 & 9.10 & 19.95  & 133670 & 916167 & 29.58 & 2.41 \\
stsb roberta base & 22.04 & 0.12 & 8.90 & 19.68  & 135544 & 925948 & 29.75 & 2.44 \\
stsb roberta large & 21.10 & 0.12 & \textbf{8.29} & 20.11  & 134706 & 961836 & 29.18 & \textbf{2.24} \\
xlm distilroberta paraphrase & 22.08 & 0.12 & 9.05 & 19.35  & 138934 & 951708 & 29.60 & 2.44 \\
\hline
    \end{tabular}
    \caption{Results from our proposed metrics plus some additional information occurring from our metrics per model}
    \label{tab:our_metrics}
\end{table*}



\subsection{Adversarial re-ranking evaluation} 

Adjectives were substituted by their antonyms in the 30.93\% of total queries, while color and size substitutions occurred in the 47.04\% and 30.77\% of queries respectively, producing $q^*_i\neq q_i$. Updated query representations resulted in re-ranking of instances; specifically, on average almost 70\% of instances changed position in $R^*$ comparing to $R$ as presented in Table \ref{tab:adversarial-changes}. \textit{Adv.query} column refers to number of perturbed queries, while \textit{Lower, Higher, Same} columns refer to the position change.

\begin{table}[b!]
\centering
\begin{tabular}{l|c|p{22pt}p{21pt}p{22pt}p{22pt}}
\hline
Adv.   &  Adv.   &     &  $\mathsf{rank}^*$  & \% \\
Change &  query  & Lower  & Higher   & Same  \\
\hline
Antonym    & 10523  & 41.30 & 27.73 & 30.97 \\
Color-all  & 16007  & 48.29 & 24.04 & 27.68 \\
Color-in   & 16007  & 49.73 & 22.35 & 27.92 \\
Size       & 10471  & 37.26 & 28.28 & 34.46 \\
\hline
\end{tabular}
\caption{Changes for all adversarial perturbations.}
\label{tab:adversarial-changes}
\end{table}

By qualitatively assessing adversarial failures, we observe that adversarially perturbed semantics are rather bypassed in favor of preserving object class. Even if this could imply representation robustness, on the other hand it can be attributed to language model biases towards object identities. In any case, existing ranking metrics cannot indicate potential biases, patterns and rules in the linguistic representations due to their opaque nature.

\subsection{Non-explainable evaluation vulnerabilities} 

Overall,  despite the re-arrangements of individual instances, $R^*$ was only marginally altered in global level for any of the adversarial perturbations according to all \textit{query-agnostic} metrics (Table \ref{tab:adversarial-all}). Therefore, either by providing meaningful and relevant queries or conceptually divergent ones, the response of a semantic similarity system is virtually the same.
This invariance over non-minimal interventions generally questions the trustworthiness of opaque ranking metrics, highlighting even more the need for explainable evaluation. 
Even then, query semantics are not taken into account, generally exposing query-agnostic evaluation against adversarial attacks: even if the best possible answer to a query is returned based on similarity measures, how can we ensure that it is good enough in terms of actual relevance? The query-corpus relevance can be easily and explicitly measured via their common concepts, an approach followed in \textit{query-informed} evaluation. To this end, we conclude that \textit{explainable} and, even better,  \textit{query-informed}  metrics are necessary to ensure evaluation robustness.


\begin{table}[h!]
\centering
\begin{tabular}{l|p{18pt}p{20pt}|p{18pt}p{20pt}|c}
\hline
Adv.   &  Recall   & (\%)  & MRR  & (\%)   & Fails \\
Change & $@1$  & $@10$  & $@10$ &  $@all$   & (\%)\\
\hline
Original & 15.31  & 39.48  & 22.11 & 23.42  & 84.69 \\
Antonym & 15.13  & 38.82  & 21.76 & 23.07  & 84.87 \\
Color-all & 14.52  & 38.16  & 21.15 & 22.48  & 85.48 \\
Color-in & 14.52 & 38.05  & 21.12 & 22.46  & 85.48 \\
Size & 15.19  & 39.32  & 21.97 & 23.29  & 84.81 \\
\hline
\end{tabular}
\caption{Rank results on adversarial queries.}
\label{tab:adversarial-all}
\end{table}

\section{Conclusion}
In this work, we presented an evaluation framework for text-image retrieval and experimented with pre-trained transformer-based semantic similarity models. Our approach achieved in capturing representation patterns and evaluation shortcomings of widely used metrics in local and global level. As future work, we aspire to extend our automated metrics to include attributes and spatial relationships between concepts, and produce adversarial re-rankings using verb antonyms, singular-plural sentence transformations and rare synonyms of salient concepts.

\bibliography{anthology,custom}
\bibliographystyle{acl_natbib}

\appendix

\section{Appendix} 
\label{sec:appendix}

\subsection{Qualitative results of retrieved vs ground truth pairs} 
\begin{figure}[h!]
\begin{subfigure}[b]{0.47\columnwidth}
\centering 
\includegraphics[width=\columnwidth, height=2.5cm]{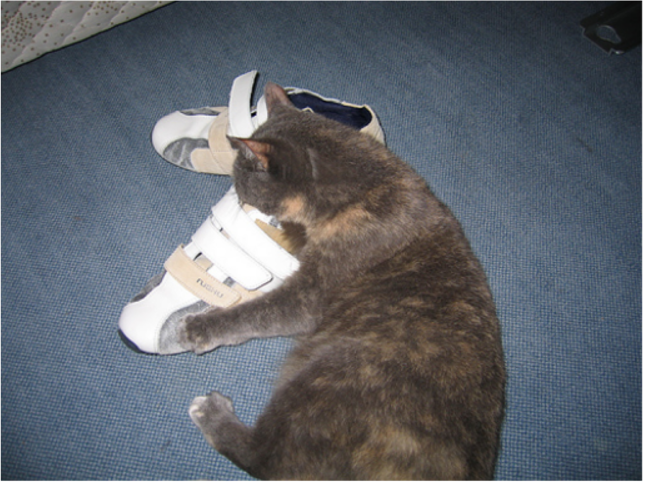}
\end{subfigure}\hfill
\begin{subfigure}[b]{0.47\columnwidth}
\centering 
\includegraphics[width=\columnwidth,height=2.5cm]{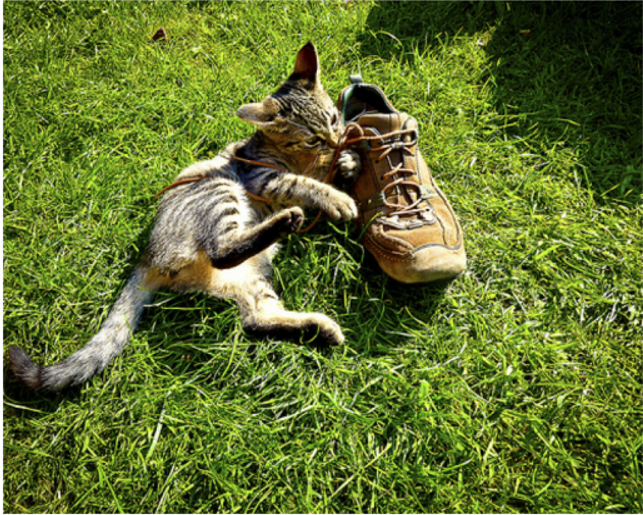}
\end{subfigure}
\caption{\textbf{\textit{Detailed semantics.}}A cat playing with a shoe in a grassy field.}
\label{fig:detail}
\end{figure}

The following Figures \ref{fig:alternatives}, \ref{fig:color}, \ref{fig:enumerate}, \ref{fig:detail} demonstrate some interesting results regarding retrieved images and their ground truth matchings with respect to a given query. Images to the left correspond to the retrieved image $I_{r_i}$, while images to the right denote the ground truth image $I_{g_i}$ with respect to a query $q_i$ appearing in the caption. The caption also mentions the failure category the images belong, according to human evaluation results.
\begin{figure}[h!]
\begin{subfigure}[b]{0.47\columnwidth}
\centering 
\includegraphics[width=\columnwidth, height=2.5cm]{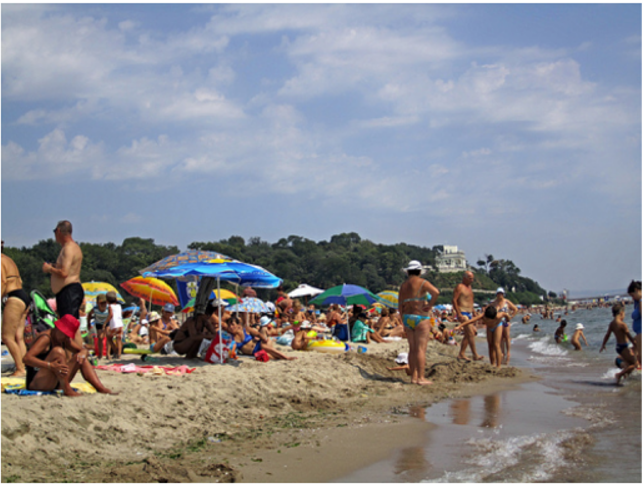}
\end{subfigure}\hfill
\begin{subfigure}[b]{0.47\columnwidth}
\centering 
\includegraphics[width=\columnwidth,height=2.5cm]{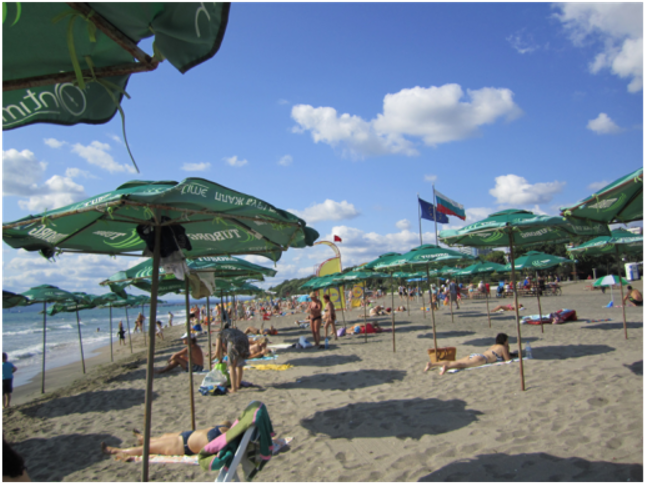}
\end{subfigure}
\caption{\textbf{\textit{Successful alternative.}}Many people are relaxing under their umbrellas on the beach.}
\label{fig:alternatives}
\end{figure}

\begin{figure}[h!]
\begin{subfigure}[b]{0.47\columnwidth}
\centering 
\includegraphics[width=\columnwidth, height=3.5cm]{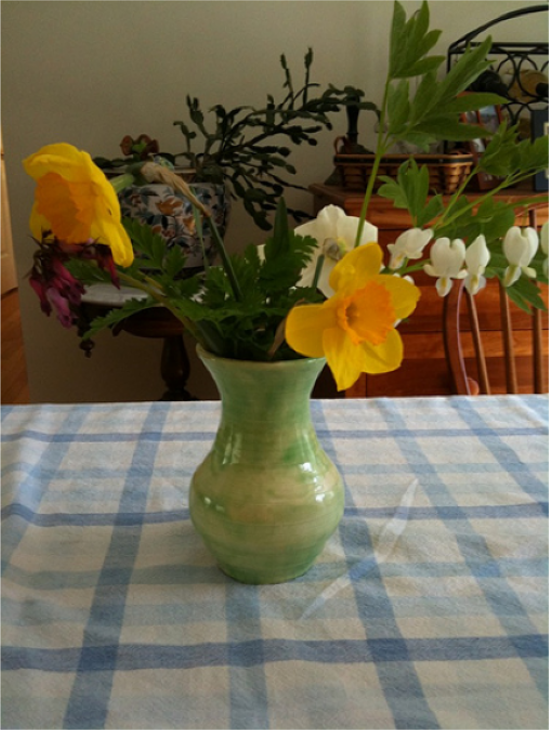}
\end{subfigure}\hfill
\begin{subfigure}[b]{0.47\columnwidth}
\centering 
\includegraphics[width=\columnwidth,height=3.5cm]{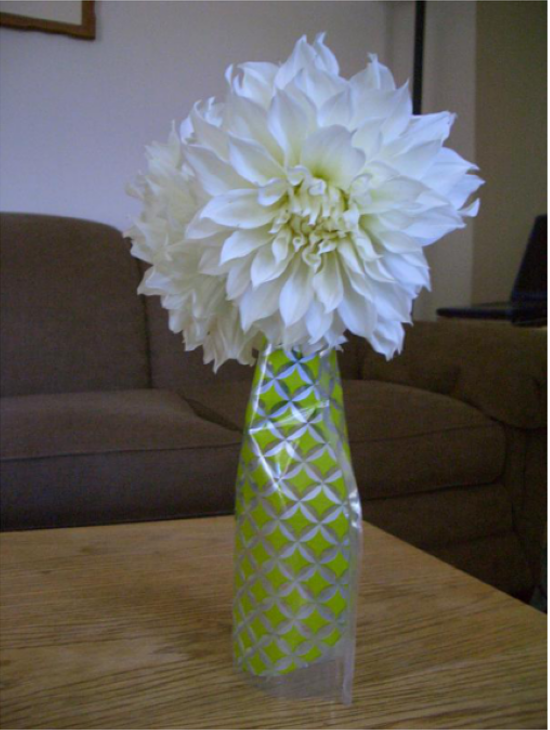}
\end{subfigure}
\caption{\textbf{\textit{Object color.}}A vase sitting on a table with white flowers in it.}
\label{fig:color}
\end{figure}

\begin{figure}[h!]
\begin{subfigure}[b]{0.47\columnwidth}
\centering 
\includegraphics[width=\columnwidth, height=2.5cm]{image_pairs/other-enum1.PNG}
\end{subfigure}\hfill
\begin{subfigure}[b]{0.47\columnwidth}
\centering 
\includegraphics[width=\columnwidth,height=2.5cm]{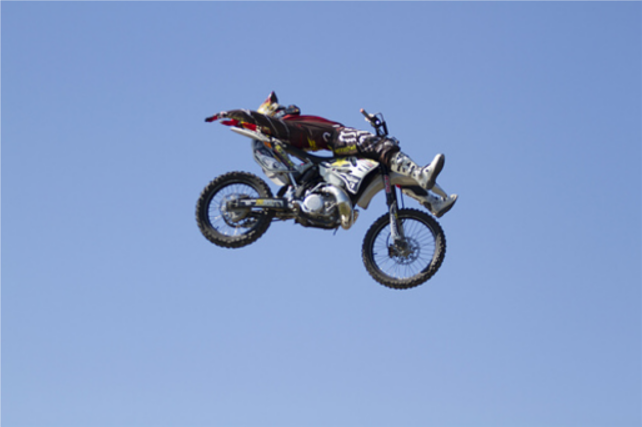}
\end{subfigure}
\caption{\textbf{\textit{Object enumeration.}}A dirt bike rider performing a stunt while in the air.}
\label{fig:enumerate}
\end{figure}

\subsubsection{Human evaluation details}
\begin{figure}[!h]
    \centering
        \includegraphics[width=\columnwidth]{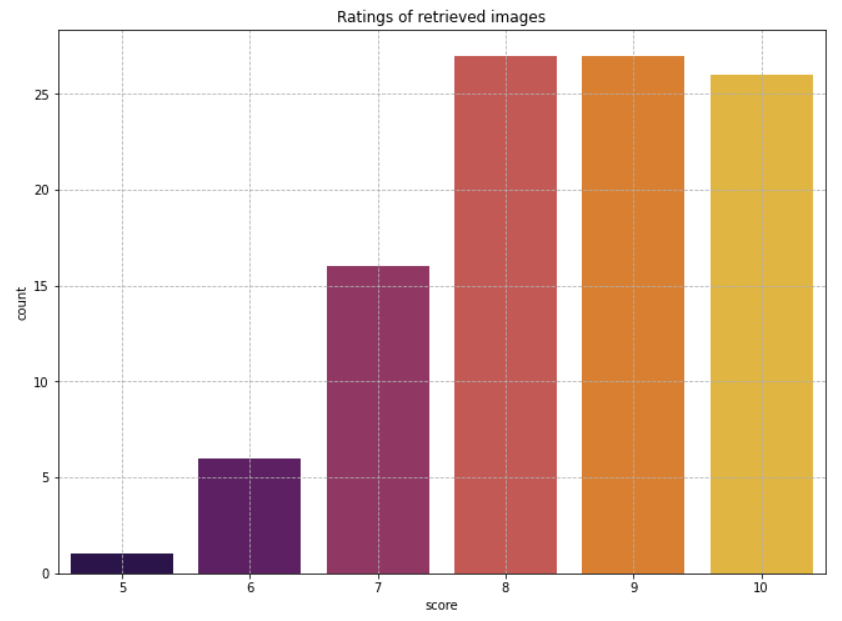}
        \caption{Human evaluation ratings (1-10).}
    \label{fig:human_rates}
\end{figure}

\begin{figure}[!h]
    \centering
        \includegraphics[width=\columnwidth]{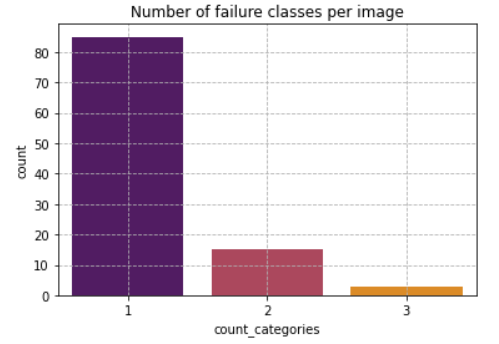}
        \caption{Number of marked disagreeing semantics per $I_{g_i}, I_{r_i}$ pair for all evaluated image pairs in $\mathcal{F}$.}
    \label{fig:semantics}
\end{figure}

We present some distributions regarding human evaluation experiments. Figure \ref{fig:human_rates} regards the rating distribution according to our evaluators' perception of ground truth-retrieved image relevance with respect to the given query. Figure \ref{fig:semantics} presents the number of failed semantics categories per image.

\subsection{Analyzing adversarial position changes}
\paragraph{Antonym-based queries}
Substituting adjectives with their antonyms was applicable on 10523 queries which resulted in updated embedding representations: the cosine similarity $cos(u^{q}_i,u^{q*}_i)<1$. By exclusively considering adversarial instances with updated representations $u^{q}_i \neq u^{q*}_i$, we observed that 
4346  instances (41.30\%) were ranked lower than the original ones,
2918  instances (27.73\%) were ranked higher, and
3259  instances (30.97\%) remained in the same position. 

\paragraph{Size-based queries} 10471 queries of the dataset where perturbed with respect to size-oriented words. Consequently,
3902  instances (37.26\%) were ranked lower than the original ones,
2961 instances (28.28\%) were ranked higher, and
3608 instances (34.46\%) remained in the same position.

\paragraph{Color-based queries}
16007 queries that contain colors were adversarially perturbed.
Regarding the \textit{color-in}) experiment and by considering adversarial instances with updated representations, 
7960  instances (49.73\%) were ranked lower, 
3578  instances (22.35\%) were ranked higher, and 
4469  instances (27.92\%) remained in the same position. 
In the \textit{color-all} experiment, we observe that 
7729 instances (48.29\%) were ranked lower than the original ones, 
3848 instances (24.04\%) were ranked higher, and 
4430 instances (27.68\%) remained in the same position.

Those results are summarized in Table \ref{tab:adversarial-changes}.


\subsection{Ranking results}
In this section we present all ranking experiments conducted in this work. Those include all 3 datasets (Flickr8K, Flickr30K and VG$\,\cap\,$COCO) and representation choices (summarizing with T5 \cite{t5} or pegasus \cite{pegasus} abstractive summarizers before using an SBERT model, or embed corpus sentences intependently using an SBERT model and then calculate the averaged embedding representation). 

Tables \ref{tab:8k}, \ref{tab:8k-t5}, \ref{tab:8k-pegasus} refer to Flickr8K experiments; Tables \ref{tab:30k}, \ref{tab:30k-t5} refer to Flickr30k experiments; finally, \ref{tab:vgcoco}, \ref{tab:vgcoco-t5}, \ref{tab:vgcoco-pegasus} refer to VG$\,\cap\,$COCO experiments.

\begin{table*}[!h]
\centering
\begin{tabular}{l|p{25pt}p{25pt}p{25pt}|p{25pt}p{25pt}p{25pt}|c|c}
\hline
Name     &  & Recall(\%)$\uparrow$ &  &  & MRR(\%)$\uparrow$ &   & Median & Fails \\
     & $@1$ & $@5$ & $@10$ & $@5$ & $@10$ &  $@all$ & Rank $\downarrow$ & (\%) $\downarrow$  \\
\hline
all MiniLM L12 & 0.4486 & 0.6670 & 0.7477 & 0.5307 & 0.5418 & 0.5504 & 2 & 55.14 \\
all MiniLM L6 & 0.4354 & 0.6568 & 0.7392 & 0.5183 & 0.5294 & 0.5380 & 2 & 56.46 \\
all distilroberta & 0.4699 & 0.6972 & 0.7754 & 0.5557 & 0.5662 & 0.5744 & 2 & 53.01 \\
all mpnet base & 0.4664 & 0.6872 & 0.7669 & 0.5503 & 0.5610 & 0.5692 & 2 & 53.36 \\
all roberta large & 0.4858 & 0.7081 & 0.7900 & 0.5700 & 0.5811 & 0.5886 & 2 & 51.42 \\
multi qa MiniLM L6 cos & 0.3662 & 0.5687 & 0.6541 & 0.4411 & 0.4526 & 0.4628 & 3 & 63.38 \\
multi qa distilbert cos & 0.3797 & 0.5874 & 0.6738 & 0.4583 & 0.4699 & 0.4795 & 3 & 62.03 \\
multi qa mpnet base cos & 0.3990 & 0.6097 & 0.6990 & 0.4784 & 0.4904 & 0.4996 & 3 & 60.10 \\
nli distilroberta base & 0.4011 & 0.6256 & 0.7152 & 0.4864 & 0.4984 & 0.5076 & 2 & 59.89 \\
nq distilbert base & 0.3260 & 0.5363 & 0.6187 & 0.4037 & 0.4149 & 0.4255 & 4 & 67.40 \\
paraphrase MiniLM L12 & 0.5289 & 0.7602 & 0.8338 & 0.6173 & 0.6273 & 0.6337 & \textbf{1} & 47.11 \\
paraphrase MiniLM L3 & 0.5064 & 0.7337 & 0.8183 & 0.5933 & 0.6047 & 0.6115 & \textbf{1} & 49.36 \\
paraphrase MiniLM L6 & 0.5138 & 0.7497 & 0.8259 & 0.6044 & 0.6147 & 0.6212 & \textbf{1} & 48.62 \\
paraphrase TinyBERT L6 & \textbf{0.5685} & \textbf{0.8060} & \textbf{0.8758} & \textbf{0.6604} & \textbf{0.6700} & \textbf{0.6749} & \textbf{1} & \textbf{43.15} \\
paraphrase albert base & 0.4776 & 0.7167 & 0.7957 & 0.5684 & 0.5789 & 0.5864 & 2 & 52.24 \\
paraphrase distilroberta & 0.5296 & 0.7643 & 0.8399 & 0.6197 & 0.6298 & 0.6361 & \textbf{1} & 47.04 \\
paraphrase mpnet & 0.4936 & 0.7318 & 0.8118 & 0.5846 & 0.5953 & 0.6022 & 2 & 50.64 \\
stsb distilroberta base & 0.4106 & 0.6219 & 0.7044 & 0.4902 & 0.5013 & 0.5109 & 2 & 58.94 \\
stsb mpnet base & 0.4334 & 0.6517 & 0.7359 & 0.5159 & 0.5272 & 0.5356 & 2 & 56.66 \\
stsb roberta base & 0.4206 & 0.6354 & 0.7201 & 0.5017 & 0.5131 & 0.5224 & 2 & 57.94 \\
stsb roberta large & 0.3678 & 0.5694 & 0.6575 & 0.4426 & 0.4545 & 0.4643 & 3 & 63.22 \\
xlm distilroberta paraphrase & 0.4138 & 0.6382 & 0.7255 & 0.4979 & 0.5096 & 0.5187 & 2 & 58.62 \\
\hline
\end{tabular}
\caption{Rank results for Flickr8k. Bold entries indicate best results.}
\label{tab:8k}
\end{table*}

\begin{table*}[!h]
\centering
\begin{tabular}{l|p{25pt}p{25pt}p{25pt}|p{25pt}p{25pt}p{25pt}|c|c}
\hline
Name     &  & Recall(\%)$\uparrow$ &  &  & MRR(\%)$\uparrow$ &   & Median & Fails \\
     & $@1$ & $@5$ & $@10$ & $@5$ & $@10$ &  $@all$ & Rank $\downarrow$ & (\%) $\downarrow$  \\
\hline
all MiniLM L12 & 0.3139 & 0.5285 & 0.6213 & 0.3932 & 0.4056 & 0.4169 & 5 & 68.61 \\
all MiniLM L6 & 0.3100 & 0.5193 & 0.6143 & 0.3870 & 0.3996 & 0.4106 & 5 & 69.00 \\
all distilroberta & 0.3421 & 0.5717 & 0.6600 & 0.4281 & 0.4400 & 0.4508 & 4 & 65.79 \\
all mpnet base & 0.3242 & 0.5410 & 0.6347 & 0.4051 & 0.4176 & 0.4285 & 4 & 67.58 \\
all roberta large & 0.3548 & 0.5818 & 0.6787 & 0.4391 & 0.4521 & 0.4624 & 3 & 64.52 \\
multi qa MiniLM L6 cos & 0.2262 & 0.3965 & 0.4789 & 0.2887 & 0.2996 & 0.3115 & 12 & 77.38 \\
multi qa distilbert cos & 0.2351 & 0.4092 & 0.4928 & 0.2993 & 0.3104 & 0.3220 & 11 & 76.49 \\
multi qa mpnet base cos & 0.2406 & 0.4306 & 0.5197 & 0.3103 & 0.3220 & 0.3337 & 9 & 75.94 \\
nli distilroberta base & 0.2856 & 0.4928 & 0.5852 & 0.3626 & 0.3748 & 0.3864 & 6 & 71.44 \\
nq distilbert base & 0.1646 & 0.3058 & 0.3766 & 0.2165 & 0.2258 & 0.2376 & 28 & 83.54 \\
paraphrase MiniLM L12 & 0.4071 & 0.6581 & 0.7492 & 0.5008 & 0.5131 & 0.5222 & \textbf{2} & 59.29 \\
paraphrase MiniLM L3 & 0.3864 & 0.6281 & 0.7266 & 0.4772 & 0.4904 & 0.4999 & 3 & 61.36 \\
paraphrase MiniLM L6 & 0.3960 & 0.6470 & 0.7391 & 0.4902 & 0.5024 & 0.5119 & \textbf{2} & 60.40 \\
paraphrase TinyBERT L6 & \textbf{0.4390} & \textbf{0.6995} & \textbf{0.7941} & \textbf{0.5376} & \textbf{0.5505} & \textbf{0.5584} & \textbf{2} & \textbf{56.10} \\
paraphrase albert base & 0.3618 & 0.6014 & 0.6937 & 0.4517 & 0.4642 & 0.4745 & 3 & 63.82 \\
paraphrase albert small & 0.4110 & 0.6699 & 0.7594 & 0.5083 & 0.5207 & 0.5300 & \textbf{2} & 58.90 \\
paraphrase distilroberta & 0.4085 & 0.6667 & 0.7613 & 0.5042 & 0.5169 & 0.5253 & \textbf{2} & 59.15 \\
paraphrase mpnet & 0.3856 & 0.6233 & 0.7186 & 0.4752 & 0.4881 & 0.4976 & 3 & 61.44 \\
stsb distilroberta base & 0.2886 & 0.4881 & 0.5746 & 0.3625 & 0.3741 & 0.3855 & 6 & 71.14 \\
stsb mpnet base & 0.3105 & 0.5185 & 0.6043 & 0.3879 & 0.3995 & 0.4106 & 5 & 68.95 \\
stsb roberta base & 0.2971 & 0.5008 & 0.5882 & 0.3725 & 0.3843 & 0.3957 & 5 & 70.29 \\
stsb roberta large & 0.2488 & 0.4173 & 0.5044 & 0.3110 & 0.3226 & 0.3346 & 10 & 75.12 \\
xlm distilroberta paraphrase & 0.2814 & 0.4899 & 0.5857 & 0.3579 & 0.3708 & 0.3821 & 6 & 71.86 \\
\hline
\end{tabular}
\caption{Rank results for Flickr8k - T5 \cite{t5} summarizer. Bold entries indicate best results.}
\label{tab:8k-t5}
\end{table*}

\begin{table*}[!h]
\centering
\begin{tabular}{l|p{25pt}p{25pt}p{25pt}|p{25pt}p{25pt}p{25pt}|c|c}
\hline
Name     &  & Recall(\%)$\uparrow$ &  &  & MRR(\%)$\uparrow$ &   & Median & Fails \\
     & $@1$ & $@5$ & $@10$ & $@5$ & $@10$ &  $@all$ & Rank $\downarrow$ & (\%) $\downarrow$  \\
\hline
all MiniLM L12 & 0.2396 & 0.4336 & 0.5203 & 0.3097 & 0.3210 & 0.3328 & 9 & 76.04 \\
all MiniLM L6 & 0.2330 & 0.4190 & 0.5135 & 0.3016 & 0.3133 & 0.3267 & 9 & 76.70 \\
all distilroberta & 0.2605 & 0.4645 & 0.5556 & 0.3359 & 0.3475 & 0.3602 & 7 & 73.95 \\
all mpnet base & 0.2519 & 0.4479 & 0.5429 & 0.3239 & 0.3369 & 0.3498 & 8 & 74.81 \\
all roberta large & 0.2686 & 0.4789 & 0.5695 & 0.3457 & 0.3576 & 0.3708 & 6 & 73.14 \\
multi qa MiniLM L6 cos & 0.1754 & 0.3163 & 0.3908 & 0.2258 & 0.2362 & 0.2482 & 25 & 82.46 \\
multi qa distilbert cos & 0.1856 & 0.3353 & 0.4126 & 0.2402 & 0.2501 & 0.2624 & 21 & 81.44 \\
multi qa mpnet base cos & 0.2008 & 0.3569 & 0.4363 & 0.2572 & 0.2681 & 0.2794 & 18 & 79.92 \\
nli distilroberta base & 0.2221 & 0.4070 & 0.4964 & 0.2888 & 0.3007 & 0.3143 & 11 & 77.79 \\
nq distilbert base & 0.1456 & 0.2712 & 0.3359 & 0.1905 & 0.1993 & 0.2110 & 40 & 85.44 \\
paraphrase MiniLM L12 & 0.3061 & 0.5358 & 0.6390 & 0.3909 & 0.4041 & 0.4156 & \textbf{4} & 69.39 \\
paraphrase MiniLM L3 & 0.2876 & 0.5024 & 0.6038 & 0.3663 & 0.3800 & 0.3930 & 5 & 71.24 \\
paraphrase MiniLM L6 & 0.2960 & 0.5234 & 0.6233 & 0.3804 & 0.3928 & 0.4055 & 5 & 70.40 \\
paraphrase TinyBERT L6 & \textbf{0.3223} & \textbf{0.5667} & \textbf{0.6696} & \textbf{0.4126} & \textbf{0.4262} & \textbf{0.4381} & \textbf{4} & \textbf{67.77} \\
paraphrase albert base & 0.2830 & 0.4909 & 0.5925 & 0.3596 & 0.3731 & 0.3856 & 6 & 71.70 \\
paraphrase albert small & 0.3071 & 0.5387 & 0.6398 & 0.3922 & 0.4058 & 0.4185 & \textbf{4} & 69.29 \\
paraphrase distilroberta & 0.3082 & 0.5404 & 0.6417 & 0.3938 & 0.4071 & 0.4185 & \textbf{4} & 69.18 \\
paraphrase mpnet & 0.2903 & 0.5055 & 0.6077 & 0.3701 & 0.3840 & 0.3964 & 5 & 70.97 \\
stsb distilroberta base & 0.2249 & 0.4028 & 0.4875 & 0.2891 & 0.3006 & 0.3132 & 12 & 77.51 \\
stsb mpnet base & 0.2369 & 0.4217 & 0.5086 & 0.3049 & 0.3161 & 0.3279 & 10 & 76.31 \\
stsb roberta base & 0.2298 & 0.4101 & 0.4989 & 0.2948 & 0.3063 & 0.3194 & 11 & 77.02 \\
stsb roberta large & 0.1884 & 0.3359 & 0.4144 & 0.2420 & 0.2524 & 0.2648 & 20 & 81.16 \\
xlm distilroberta paraphrase & 0.2231 & 0.4032 & 0.4967 & 0.2880 & 0.3008 & 0.3142 & 11 & 77.69 \\
\hline
\end{tabular}
\caption{Rank results for Flickr8k - Pegasus \cite{pegasus} summarizer. Bold entries indicate best results.}
\label{tab:8k-pegasus}
\end{table*}

\begin{table*}[h!]
\centering
\begin{tabular}{l|p{25pt}p{25pt}p{25pt}|p{25pt}p{25pt}p{25pt}|c|c}
\hline
Name     &  & Recall(\%)$\uparrow$ &  &  & MRR(\%)$\uparrow$ &   & Median & Fails \\
     & $@1$ & $@5$ & $@10$ & $@5$ & $@10$ &  $@all$ & Rank $\downarrow$ & (\%) $\downarrow$  \\
\hline
all MiniLM L12 & 0.3076 & 0.5155 & 0.6063 & 0.3848 & 0.3969 & 0.4075 & 5 & 69.24 \\
all MiniLM L6 & 0.2961 & 0.5021 & 0.5924 & 0.3727 & 0.3849 & 0.3956 & 5 & 70.39 \\
all distilroberta & 0.3276 & 0.5439 & 0.6357 & 0.4082 & 0.4204 & 0.4310 & 4 & 67.24 \\
all mpnet base & 0.3233 & 0.5372 & 0.6304 & 0.4030 & 0.4155 & 0.4259 & 4 & 67.67 \\
all roberta large & 0.3381 & 0.5633 & 0.6527 & 0.4223 & 0.4343 & 0.4445 & 4 & 66.19 \\
multi qa MiniLM L6 cos & 0.2397 & 0.4205 & 0.5042 & 0.3058 & 0.3169 & 0.3281 & 10 & 66.19 \\
multi qa distilbert cos & 0.2530 & 0.4379 & 0.5252 & 0.3210 & 0.3327 & 0.3440 & 9 & 74.70 \\
multi qa mpnet base cos & 0.2730 & 0.4683 & 0.5555 & 0.3453 & 0.3570 & 0.3681 & 7 & 72.70 \\
nli distilroberta base & 0.2770 & 0.4780 & 0.5718 & 0.3512 & 0.3638 & 0.3750 & 6 & 72.30 \\
nq distilbert base & 0.2131 & 0.3841 & 0.4653 & 0.2760 & 0.2868 & 0.2983 & 14 & 78.69 \\
paraphrase MiniLM L12 & 0.3687 & 0.6148 & 0.7074 & 0.4606 & 0.4731 & 0.4831 & 3 & 63.13 \\
paraphrase MiniLM L3 & 0.3437 & 0.5826 & 0.6764 & 0.4328 & 0.4453 & 0.4559 & 3 & 65.63 \\
paraphrase MiniLM L6 & 0.3526 & 0.5938 & 0.6888 & 0.4425 & 0.4553 & 0.4657 & 3 & 64.74 \\
paraphrase TinyBERT L6 & \textbf{0.4199} & \textbf{0.6844} & \textbf{0.7850} & \textbf{0.5198} & \textbf{0.5333} & \textbf{0.5421} & \textbf{2} & \textbf{58.01} \\
paraphrase albert base & 0.3294 & 0.5548 & 0.6482 & 0.4132 & 0.4257 & 0.4365 & 4 & 67.06 \\
paraphrase albert small & 0.3831 & 0.6416 & 0.7432 & 0.4799 & 0.4935 & 0.5032 & 3 & 61.69 \\
paraphrase distilroberta & 0.3796 & 0.6224 & 0.7230 & 0.4705 & 0.4840 & 0.4937 & 3 & 62.04 \\
paraphrase mpnet & 0.3476 & 0.5804 & 0.6730 & 0.4341 & 0.4466 & 0.4570 & 3 & 65.24 \\
stsb distilroberta base & 0.2765 & 0.4716 & 0.5626 & 0.3487 & 0.3610 & 0.3724 & 7 & 72.35 \\
stsb mpnet base & 0.2935 & 0.4958 & 0.5846 & 0.3680 & 0.3799 & 0.3909 & 6 & 70.65 \\
stsb roberta base & 0.2864 & 0.4888 & 0.5757 & 0.3612 & 0.3729 & 0.3840 & 6 & 71.36 \\
stsb roberta large & 0.2410 & 0.4207 & 0.5037 & 0.3067 & 0.3178 & 0.3292 & 10 & 75.90 \\
xlm distilroberta paraphrase & 0.2816 & 0.4832 & 0.5732 & 0.3559 & 0.3679 & 0.3790 & 6 & 71.84 \\
\hline
\end{tabular}
\caption{Rank results Flickr30k. Bold entries indicate best results.}
\label{tab:30k}
\end{table*}

\begin{table*}[h!]
\centering
\begin{tabular}{l|p{25pt}p{25pt}p{25pt}|p{25pt}p{25pt}p{25pt}|c|c}
\hline
Name     &  & Recall(\%)$\uparrow$ &  &  & MRR(\%)$\uparrow$ &   & Median & Fails \\
     & $@1$ & $@5$ & $@10$ & $@5$ & $@10$ &  $@all$ & Rank $\downarrow$ & (\%) $\downarrow$  \\
\hline
all MiniLM L12 & 0.2068 & 0.3845 & 0.4710 & 0.2721 & 0.2836 & 0.2955 & 13 & 79.32 \\
all MiniLM L6 & 0.2012 & 0.3739 & 0.4595 & 0.2645 & 0.2758 & 0.2877 & 14 & 79.88 \\
all distilroberta & 0.2242 & 0.4159 & 0.5083 & 0.2948 & 0.3072 & 0.3193 & 10 & 77.58 \\
all mpnet base & 0.2138 & 0.3975 & 0.4837 & 0.2812 & 0.2927 & 0.3047 & 12 & 78.62 \\
all roberta large & 0.2373 & 0.4321 & 0.5264 & 0.3091 & 0.3217 & 0.3337 & 9 & 76.27 \\
multi qa MiniLM L6 cos & 0.1380 & 0.2665 & 0.3357 & 0.1847 & 0.1939 & 0.2051 & 42 & 86.20 \\
multi qa distilbert cos & 0.1443 & 0.2820 & 0.3552 & 0.1942 & 0.2039 & 0.2152 & 35 & 85.57 \\
multi qa mpnet base cos & 0.1515 & 0.2976 & 0.3733 & 0.2049 & 0.2150 & 0.2267 & 29 & 84.85 \\
nli distilroberta base & 0.1856 & 0.3556 & 0.4401 & 0.2477 & 0.2590 & 0.2712 & 16 & 81.44 \\
nq distilbert base & 0.0951 & 0.1963 & 0.2530 & 0.1317 & 0.1392 & 0.1493 & 100 & 90.49 \\
paraphrase MiniLM L12 & 0.2740 & 0.4970 & 0.5988 & 0.3560 & 0.3696 & 0.3817 & 6 & 72.60 \\
paraphrase MiniLM L3 & 0.2493 & 0.4630 & 0.5649 & 0.3274 & 0.3410 & 0.3534 & 7 & 75.07 \\
paraphrase MiniLM L6 & 0.2644 & 0.4810 & 0.5825 & 0.3441 & 0.3577 & 0.3700 & 6 & 73.56 \\
paraphrase TinyBERT L6 & \textbf{0.3032} & \textbf{0.5555} & \textbf{0.6688} & \textbf{0.3958} & \textbf{0.4109} & \textbf{0.4230} & \textbf{4} & \textbf{69.68} \\
paraphrase albert base & 0.2400 & 0.4428 & 0.5389 & 0.3142 & 0.3271 & 0.3394 & 8 & 76.00 \\
paraphrase albert small & 0.2759 & 0.5140 & 0.6218 & 0.3627 & 0.3771 & 0.3897 & 5 & 72.41 \\
paraphrase distilroberta & 0.2803 & 0.5082 & 0.6119 & 0.3639 & 0.3777 & 0.3897 & 5 & 71.97 \\
paraphrase mpnet & 0.2584 & 0.4687 & 0.5664 & 0.3351 & 0.3482 & 0.3603 & 7 & 74.16 \\
stsb distilroberta base & 0.1800 & 0.3424 & 0.4263 & 0.2395 & 0.2507 & 0.2627 & 18 & 82.00 \\
stsb mpnet base & 0.1965 & 0.3724 & 0.4557 & 0.2605 & 0.2717 & 0.2836 & 15 & 80.35 \\
stsb roberta base & 0.1904 & 0.3547 & 0.4389 & 0.2503 & 0.2615 & 0.2735 & 17 & 80.96 \\
stsb roberta large & 0.1456 & 0.2811 & 0.3543 & 0.1945 & 0.2042 & 0.2159 & 34 & 85.44 \\
xlm distilroberta paraphrase & 0.1792 & 0.3467 & 0.4286 & 0.2403 & 0.2513 & 0.2635 & 18 & 82.08 \\
\hline
\end{tabular}
\caption{Rank results Flickr30k - T5 \cite{t5} summarizer. Bold entries indicate best results.}
\label{tab:30k-t5}
\end{table*}

\begin{table*}[h!]
\centering
\begin{tabular}{l|p{25pt}p{25pt}p{25pt}|p{25pt}p{25pt}p{25pt}|c|c}
\hline
Name     &  & Recall(\%)$\uparrow$ &  &  & MRR(\%)$\uparrow$ &   & Median & Fails \\
     & $@1$ & $@5$ & $@10$ & $@5$ & $@10$ &  $@all$ & Rank $\downarrow$ & (\%) $\downarrow$  \\
\hline
all MiniLM L12 & 13.31 & 26.93 & 34.31 & 18.18 & 19.16 & 20.45 & 34 & 86.69 \\
all MiniLM L6 & 13.07 & 26.80 & 34.26 & 18.02 & 19.02 & 20.29 & 35 & 86.93 \\
all distilroberta & 12.20 & 26.10 & 33.58 & 17.22 & 18.22 & 19.55 & 35 & 87.80 \\
all mpnet base & 12.64 & 26.12 & 33.82 & 17.45 & 18.47 & 19.78 & 35 & 87.36 \\
all roberta large & 12.01 & 25.63 & 33.42 & 16.87 & 17.91 & 19.24 & 35 & 87.99 \\
multi qa MiniLM L6 cos & 9.29 & 20.10 & 26.77 & 13.14 & 14.03 & 15.29 & 60 & 90.71 \\
multi qa distilbert cos & 9.22 & 20.35 & 27.04 & 13.16 & 14.04 & 15.32 & 57 & 90.78 \\
multi qa mpnet base cos & 9.53 & 21.03 & 27.81 & 13.63 & 14.52 & 15.79 & 55 & 90.47 \\
nli distilroberta base & 11.73 & 24.94 & 32.33 & 16.48 & 17.46 & 18.75 & 39 & 88.27 \\
nq distilbert base & 8.03 & 17.95 & 23.95 & 11.53 & 12.32 & 13.54 & 76 & 91.97 \\
paraphrase MiniLM L12 & 14.24 & 29.89 & 38.34 & 19.89 & 21.02 & 22.37 & 24 & 85.76 \\
paraphrase MiniLM L3 & \textbf{15.31} & \textbf{30.92} & \textbf{39.48} & \textbf{20.97} & \textbf{22.11} & \textbf{23.42} & 23 & \textbf{84.69} \\
paraphrase MiniLM L6 & 14.39 & 30.01 & 38.51 & 19.96 & 21.10 & 22.46 & 24 & 85.61 \\
paraphrase TinyBERT L6 & 14.55 & 30.38 & 39.12 & 20.27 & 21.43 & 22.82 & \textbf{22} & 85.45 \\
paraphrase albert base & 13.12 & 27.83 & 36.11 & 18.43 & 19.54 & 20.91 & 28 & 86.88 \\
paraphrase albert small & 14.56 & 30.25 & 39.04 & 20.23 & 21.40 & 22.75 & 23 & 85.44 \\
paraphrase distilroberta & 14.51 & 30.39 & 38.91 & 20.23 & 21.36 & 22.74 & \textbf{22} & 85.49 \\
paraphrase mpnet & 13.99 & 28.99 & 37.40 & 19.39 & 20.50 & 21.84 & 26 & 86.01 \\
stsb distilroberta base & 13.41 & 27.04 & 34.28 & 18.32 & 19.28 & 20.55 & 35 & 86.59 \\
stsb mpnet base & 14.05 & 28.26 & 35.93 & 19.23 & 20.24 & 21.49 & 32 & 85.95 \\
stsb roberta base & 13.69 & 27.38 & 34.79 & 18.67 & 19.65 & 20.92 & 34 & 86.31 \\
stsb roberta large & 10.32 & 22.13 & 28.71 & 14.60 & 15.47 & 16.73 & 53 & 89.68 \\
xlm distilroberta paraphrase & 11.59 & 24.62 & 31.87 & 16.26 & 17.23 & 18.52 & 40 & 88.41 \\
\hline
\end{tabular}
\caption{Rank results for VG$\,\cap\,$COCO. Bold entries indicate best results.}
\label{tab:vgcoco}
\end{table*}

\begin{table*}[h!]
\centering
\begin{tabular}{l|p{25pt}p{25pt}p{25pt}|p{25pt}p{25pt}p{25pt}|c|c}
\hline
Name     &  & Recall(\%)$\uparrow$ &  &  & MRR(\%)$\uparrow$ &   & Median & Fails \\
     & $@1$ & $@5$ & $@10$ & $@5$ & $@10$ &  $@all$ & Rank $\downarrow$ & (\%) $\downarrow$  \\
\hline
all MiniLM L12 & 0.0989 & 0.2072 & 0.2684 & 0.1378 & 0.1459 & 0.1575 & 68 & 90.11 \\
all MiniLM L6 & 0.0923 & 0.2004 & 0.2601 & 0.1313 & 0.1393 & 0.1509 & 71 & 90.77 \\
all distilroberta & 0.1034 & 0.2184 & 0.2834 & 0.1445 & 0.1532 & 0.1654 & 56 & 89.66 \\
all mpnet base & 0.0906 & 0.1969 & 0.2608 & 0.1286 & 0.1370 & 0.1488 & 69 & 90.94 \\
all roberta large & 0.1125 & 0.2340 & 0.2994 & 0.1563 & 0.1650 & 0.1769 & 52 & 88.75 \\
multi qa MiniLM L6 cos & 0.0496 & 0.1132 & 0.1558 & 0.0719 & 0.0776 & 0.0872 & 189 & 95.04 \\
multi qa distilbert cos & 0.0519 & 0.1172 & 0.1580 & 0.0748 & 0.0802 & 0.0901 & 181 & 94.81 \\
multi qa mpnet base cos & 0.0509 & 0.1233 & 0.1718 & 0.0764 & 0.0827 & 0.0931 & 148 & 94.91 \\
nli distilroberta base & 0.0914 & 0.1931 & 0.2502 & 0.1281 & 0.1357 & 0.1470 & 82 & 90.86 \\
nq distilbert base & 0.0364 & 0.0914 & 0.1285 & 0.0560 & 0.0609 & 0.0698 & 249 & 96.36 \\
paraphrase MiniLM L12 & \textbf{0.1365} & \textbf{0.2794} & \textbf{0.3579} & \textbf{0.1882} & \textbf{0.1987} & \textbf{0.2113} & \textbf{31} & \textbf{86.35} \\
paraphrase MiniLM L3 & 0.1165 & 0.2497 & 0.3239 & 0.1645 & 0.1743 & 0.1872 & 38 & 88.35 \\
paraphrase MiniLM L6 & 0.1258 & 0.2643 & 0.3386 & 0.1756 & 0.1855 & 0.1983 & 35 & 87.42 \\
paraphrase TinyBERT L6 & 0.1272 & 0.2652 & 0.3446 & 0.1767 & 0.1873 & 0.2001 & 33 & 87.28 \\
paraphrase albert base & 0.1215 & 0.2534 & 0.3263 & 0.1687 & 0.1784 & 0.1909 & 40 & 87.85 \\
paraphrase albert small & 0.1233 & 0.2572 & 0.3331 & 0.1715 & 0.1815 & 0.1942 & 37 & 87.67 \\
paraphrase distilroberta & 0.1274 & 0.2621 & 0.3355 & 0.1760 & 0.1857 & 0.1984 & 36 & 87.26 \\
paraphrase mpnet & 0.1362 & 0.2778 & 0.3552 & 0.1873 & 0.1976 & 0.2103 & \textbf{31} & 86.38 \\
stsb distilroberta base & 0.0891 & 0.1900 & 0.2450 & 0.1250 & 0.1323 & 0.1436 & 85 & 91.09 \\
stsb mpnet base & 0.0993 & 0.2094 & 0.2715 & 0.1389 & 0.1471 & 0.1589 & 66 & 90.07 \\
stsb roberta base & 0.0962 & 0.2019 & 0.2618 & 0.1339 & 0.1419 & 0.1534 & 72 & 90.38 \\
stsb roberta large & 0.0662 & 0.1506 & 0.2035 & 0.0959 & 0.1029 & 0.1141 & 109 & 93.38 \\
xlm distilroberta paraphrase & 0.0749 & 0.1735 & 0.2313 & 0.1102 & 0.1178 & 0.1292 & 90 & 92.51 \\
\hline
\end{tabular}
\caption{Rank results for VG$\,\cap\,$COCO - T5 \cite{t5} summarizer. Bold entries indicate best results.}
\label{tab:vgcoco-t5}
\end{table*}

\begin{table*}[h!]
\centering
\begin{tabular}{l|p{25pt}p{25pt}p{25pt}|p{25pt}p{25pt}p{25pt}|c|c}
\hline
Name     &  & Recall(\%)$\uparrow$ &  &  & MRR(\%)$\uparrow$ &   & Median & Fails \\
     & $@1$ & $@5$ & $@10$ & $@5$ & $@10$ &  $@all$ & Rank $\downarrow$ & (\%) $\downarrow$  \\
\hline
all MiniLM L12 & 0.0896 & 0.1898 & 0.2466 & 0.1257 & 0.1332 & 0.1439 & 91 & 91.04 \\
all MiniLM L6 & 0.0847 & 0.1829 & 0.2373 & 0.1199 & 0.1271 & 0.1380 & 95 & 91.53 \\
all distilroberta & 0.0946 & 0.2050 & 0.2660 & 0.1342 & 0.1423 & 0.1537 & 73 & 90.54 \\
all mpnet base & 0.0903 & 0.1918 & 0.2489 & 0.1267 & 0.1344 & 0.1452 & 86 & 90.97 \\
all roberta large & 0.0997 & 0.2070 & 0.2669 & 0.1383 & 0.1462 & 0.1574 & 74 & 90.03 \\
multi qa MiniLM L6 cos & 0.0490 & 0.1106 & 0.1541 & 0.0706 & 0.0764 & 0.0855 & 211 & 95.10 \\
multi qa distilbert cos & 0.0494 & 0.1136 & 0.1555 & 0.0722 & 0.0777 & 0.0869 & 208 & 95.06 \\
multi qa mpnet base cos & 0.0497 & 0.1145 & 0.1573 & 0.0722 & 0.0778 & 0.0874 & 191 & 95.03 \\
nli distilroberta base & 0.0755 & 0.1656 & 0.2194 & 0.1074 & 0.1145 & 0.1250 & 113 & 92.45 \\
nq distilbert base & 0.0436 & 0.1000 & 0.1390 & 0.0636 & 0.0688 & 0.0776 & 251 & 95.64 \\
paraphrase MiniLM L12 & \textbf{0.1154} & \textbf{0.2395} & \textbf{0.3076} & \textbf{0.1600} & \textbf{0.1690} & \textbf{0.1808} & 50 & \textbf{88.46} \\
paraphrase MiniLM L3 & 0.1030 & 0.2211 & 0.2870 & 0.1453 & 0.1540 & 0.1659 & 58 & 89.70 \\
paraphrase MiniLM L6 & 0.1103 & 0.2278 & 0.2964 & 0.1526 & 0.1617 & 0.1735 & 54 & 88.97 \\
paraphrase TinyBERT L6 & 0.1141 & 0.2373 & 0.3063 & 0.1583 & 0.1675 & 0.1796 & \textbf{48} & 88.59 \\
paraphrase albert base & 0.1003 & 0.2122 & 0.2757 & 0.1405 & 0.1490 & 0.1608 & 62 & 89.97 \\
paraphrase albert small & 0.1086 & 0.2269 & 0.2934 & 0.1512 & 0.1601 & 0.1722 & 53 & 89.14 \\
paraphrase distilroberta & 0.1140 & 0.2345 & 0.3026 & 0.1573 & 0.1664 & 0.1785 & 49 & 88.60 \\
paraphrase mpnet & 0.1099 & 0.2304 & 0.2973 & 0.1536 & 0.1625 & 0.1744 & 53 & 89.01 \\
stsb distilroberta base & 0.0751 & 0.1612 & 0.2145 & 0.1061 & 0.1131 & 0.1236 & 114 & 92.49 \\
stsb roberta base & 0.0783 & 0.1676 & 0.2211 & 0.1104 & 0.1174 & 0.1281 & 110 & 92.17 \\
stsb roberta large & 0.0586 & 0.1359 & 0.1844 & 0.0857 & 0.0921 & 0.1022 & 145 & 94.14 \\
xlm distilroberta paraphrase & 0.0706 & 0.1563 & 0.2066 & 0.1009 & 0.1075 & 0.1178 & 127 & 92.94 \\
\hline
\end{tabular}
\caption{Rank results for VG$\,\cap\,$COCO - Pegasus \cite{pegasus} summarizer. Bold entries indicate best results.}
\label{tab:vgcoco-pegasus}
\end{table*}

\end{document}